\documentclass{article}

\usepackage{microtype}
\usepackage{graphicx}
\usepackage{subfigure}
\usepackage{booktabs} % for professional tables

\usepackage{color}
\usepackage{booktabs}
\usepackage{algorithm,algorithmic}
\usepackage{sidecap, caption}
\usepackage{amsthm}
\usepackage{wrapfig}

\usepackage{hyperref}

\usepackage[accepted]{icml2023}

\usepackage{amsmath}
\usepackage{amssymb}
\usepackage{mathtools}
\usepackage{amsthm}

\usepackage[capitalize,noabbrev]{cleveref}

\theoremstyle{plain}
\newtheorem{theorem}{Theorem}[section]

\newtheorem{lemma}[theorem]{Lemma}

\theoremstyle{definition}

\newtheorem{assumption}[theorem]{Assumption}
\theoremstyle{remark}

\usepackage[textsize=tiny]{todonotes}

\icmltitlerunning{Robust Perception through Equivariance}

\begin{document}

\twocolumn[
\icmltitle{Robust Perception through Equivariance}



\icmlsetsymbol{equal}{*}

\begin{icmlauthorlist}
\icmlauthor{Chengzhi Mao}{ccc} %{equal,yyy}
% \icmlauthor{Firstname2 Lastname2}{equal,yyy,comp}
\icmlauthor{Lingyu Zhang}{ccc} %{comp}
\icmlauthor{Abhishek Vaibhav Joshi}{ccc}
\icmlauthor{Junfeng Yang}{ccc}
\icmlauthor{Hao Wang}{rrr}
\icmlauthor{Carl Vondrick}{ccc}
\end{icmlauthorlist}

\icmlaffiliation{ccc}{Department of Computer Science, Columbia University, New York, USA}
\icmlaffiliation{rrr}{Department of Computer Science, Rutgers University, New Jersey, USA}

\icmlcorrespondingauthor{Chengzhi Mao}{cm3797@columbia.edu}

\icmlkeywords{Machine Learning, ICML}

\vskip 0.3in
]



\printAffiliationsAndNotice{}  % leave blank if no need to mention equal contribution

\begin{abstract}
Deep networks for computer vision are not reliable when they encounter adversarial examples. In this paper, we introduce a framework that uses the dense intrinsic constraints in natural images to robustify inference. By introducing constraints at inference time, we can shift the burden of robustness from training to testing, thereby allowing the model to dynamically adjust to each individual image's unique and potentially novel characteristics at inference time. Our theoretical results show the importance of having dense constraints at inference time. In contrast to existing single-constraint methods, we propose to use equivariance, which naturally allows dense constraints at a fine-grained level in the feature space.  Our empirical experiments show that restoring feature equivariance at inference time defends against worst-case adversarial perturbations. The method obtains improved adversarial robustness on four datasets (ImageNet, Cityscapes, PASCAL VOC, and MS-COCO) on image recognition, semantic segmentation, and instance segmentation tasks. 
\end{abstract}

\def\Blue{\color{blue}}
\def\Purple{\color{purple}}

\def\A{\bm{A}}
\def\a{{\bf a}}
\def\B{{\bf B}}
\def\b{{\bf b}}
\def\C{{\bf C}}
\def\c{{\bf c}}
\def\D{{\bf D}}
\def\d{{\bf d}}
\def\E{{\bf E}}
\def\e{{\bf e}}
\def\f{{\bf f}}
\def\F{{\bf F}}
\def\K{{\bf K}}
\def\k{{\bf k}}
\def\L{{\bf L}}
\def\H{{\bf H}}
\def\h{{\bf h}}
\def\G{{\bf G}}
\def\g{{\bf g}}
\def\I{{\bf I}}
\def\J{{\bf J}}
\def\R{{\bf R}}
\def\X{{\bf X}}
\def\Y{{\bf Y}}
\def\OO{{\bf O}}
\def\oo{{\bf o}}
\def\P{{\bf P}}
\def\p{{\bf p}}
\def\Q{{\bf Q}}
\def\q{{\bf q}}
\def\r{{\bf r}}
\def\s{{\bf s}}
\def\S{{\bf S}}
\def\t{{\bf t}}
\def\T{{\bf T}}
\def\x{{\bf x}}
\def\y{{\bf y}}
\def\z{{\bf z}}
\def\Z{{\bf Z}}
\def\M{{\bf M}}
\def\m{{\bf m}}
\def\n{{\bf n}}
\def\U{{\bf U}}
\def\u{{\bf u}}
\def\V{{\bf V}}
\def\v{{\bf v}}
\def\W{{\bf W}}
\def\w{{\bf w}}
\def\0{{\bf 0}}
\def\1{{\bf 1}}

\def\AM{{\mathcal A}}
\def\EM{{\mathcal E}}
\def\FM{{\mathcal F}}
\def\TM{{\mathcal T}}
\def\UM{{\mathcal U}}
\def\XM{{\mathcal X}}
\def\YM{{\mathcal Y}}
\def\NM{{\mathcal N}}
\def\OM{{\mathcal O}}
\def\IM{{\mathcal I}}
\def\GM{{\mathcal G}}
\def\PM{{\mathcal P}}
\def\LM{{\mathcal L}}
\def\MM{{\mathcal M}}
\def\DM{{\mathcal D}}
\def\SM{{\mathcal S}}
\def\ZM{{\mathcal Z}}
\def\RB{{\mathbb R}}
\def\EB{{\mathbb E}}
\def\VB{{\mathbb V}}

\def\tx{\tilde{\bf x}}
\def\ty{\tilde{\bf y}}
\def\tz{\tilde{\bf z}}
\def\hd{\hat{d}}
\def\HD{\hat{\bf D}}
\def\hx{\hat{\bf x}}
\def\hR{\hat{R}}

\def\alp{\mbox{\boldmath$\alpha$\unboldmath}}
\def\Ome{\mbox{\boldmath$\omega$\unboldmath}}
\def\Om{\mbox{\boldmath$\Omega$\unboldmath}}
\def\bet{\mbox{\boldmath$\beta$\unboldmath}}
\def\et{\mbox{\boldmath$\eta$\unboldmath}}
\def\ep{\mbox{\boldmath$\epsilon$\unboldmath}}
\def\ph{\mbox{\boldmath$\phi$\unboldmath}}
\def\Pii{\mbox{\boldmath$\Pi$\unboldmath}}
\def\pii{\mbox{\boldmath$\pi$\unboldmath}}
\def\Ph{\mbox{\boldmath$\Phi$\unboldmath}}
\def\Ps{\mbox{\boldmath$\Psi$\unboldmath}}
\def\tha{\mbox{\boldmath$\theta$\unboldmath}}
\def\Tha{\mbox{\boldmath$\Theta$\unboldmath}}
\def\muu{\mbox{\boldmath$\mu$\unboldmath}}
\def\Si{\mbox{\boldmath$\Sigma$\unboldmath}}
\def\si{\mbox{\boldmath$\sigma$\unboldmath}}
\def\Gam{\mbox{\boldmath$\Gamma$\unboldmath}}
\def\gamm{\mbox{\boldmath$\gamma$\unboldmath}}
\def\Lam{\mbox{\boldmath$\Lambda$\unboldmath}}
\def\De{\mbox{\boldmath$\Delta$\unboldmath}}
\def\vps{\mbox{\boldmath$\varepsilon$\unboldmath}}
\def\Up{\mbox{\boldmath$\Upsilon$\unboldmath}}
\def\xii{\mbox{\boldmath$\xi$\unboldmath}}
\def\Xii{\mbox{\boldmath$\Xi$\unboldmath}}
\def\Lap{\mbox{\boldmath$\LM$\unboldmath}}
\newcommand{\ti}[1]{\tilde{#1}}

\def\tr{\mathrm{tr}}
\def\etr{\mathrm{etr}}
\def\etal{{\em et al.\/}\,}
\newcommand{\indep}{{\;\bot\!\!\!\!\!\!\bot\;}}
\def\argmax{\mathop{\rm argmax}}
\def\argmin{\mathop{\rm argmin}}
\def\vec{\text{vec}}
\def\cov{\text{cov}}
\def\dg{\text{diag}}

% \newtheorem{observation}{\textbf{Observation}}
% \newtheorem{remark}{Remark}
% \newtheorem{theorem}{Theorem}
% \newtheorem{lemma}{Lemma}
% \newtheorem{definition}{Definition}
% \newtheorem{problem}{Problem}
% \newtheorem{proposition}{Proposition}
% \newtheorem{cor}{Corollary}
% \numberwithin{theorem}{section}
% \numberwithin{lemma}{section}
% \numberwithin{remark}{section}
% \numberwithin{cor}{section}
% \numberwithin{proposition}{section}

% \newtheorem{assumption}{Assumption}

\newcommand{\tabref}[1]{Table~\ref{#1}}
\newcommand{\lemref}[1]{Lemma~\ref{#1}}
\newcommand{\thmref}[1]{Theorem~\ref{#1}}
\newcommand{\clmref}[1]{Claim~\ref{#1}}
\newcommand{\crlref}[1]{Corollary~\ref{#1}}
\newcommand{\asuref}[1]{Assumption~\ref{#1}}
\newcommand{\eqnref}[1]{Eqn.~\ref{#1}}
\newcommand{\algref}[1]{Alg.~\ref{#1}}

\renewcommand{\tilde}{\widetilde}
\renewcommand{\hat}{\widehat}
\renewcommand{\frac}{\tfrac}

\section{Introduction}

% set up prolbme. why important
% why it is hard, how others have tried
% our innovation. equi idea. Can follow iccv line.

Despite the strong performance of deep networks on computer vision benchmarks \cite{he2016deep,he2017mask, yu2017dilated}, state-of-the-art systems are not reliable when evaluated in open-world settings \cite{stylized-imgnet, hendrycks2021many-ImgR, szegedy2013intriguing, imgnet-C, AA, CW}. However, the robustness against a large number of adversarial cases remains a prerequisite necessary to deploy models in real-world applications, such as in medical imaging, healthcare, and robotics. 

% In practice, computer vision models will encounter out-of-distribution images (e.g.\ new styles), corruptions to the image (e.g.\ rain on the lens), and worst-case inputs (e.g.\ adversarial examples).

Due to the importance of this problem, there has been a large number of investigations aiming to improve the training algorithm to establish reliability. For example, data augmentation \cite{yun2019cutmix, hendrycks2021many-ImgR} and adversarial training~\cite{madry2017towards, unlabeled} improve robustness by training the model on anticipated distribution shifts and worst-case images. However, placing the burden of robustness on the training algorithm means that the model can only be robust to the corruptions that are anticipated ahead of time, which is an unrealistic assumption in an open world. In addition, retraining the model on new distributions each time can be expensive.

\begin{figure}[t]
    \centering
    \includegraphics[width=0.4\textwidth]{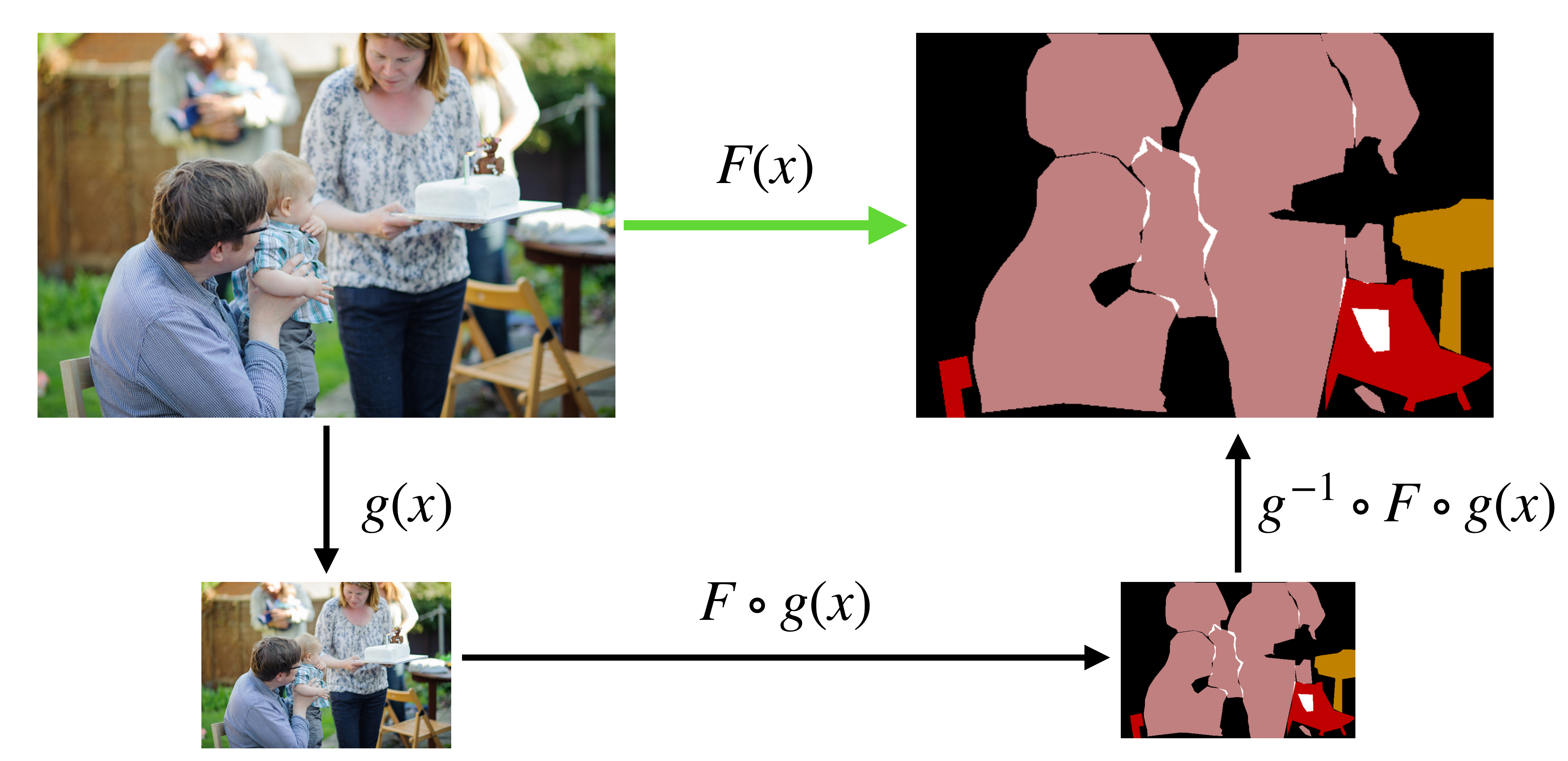}
    \caption{\small{Equivariance is shared across the input images (left) and the output labels (right), providing a dense constraint. The predictions from a model $F(x)$ should be identical to performing a spatial transformation on $x$, a forward pass of $F$, and undoing that spatial transformation on the output space (black).}}
    % \vspace{-5mm}
    \label{fig:equi}
\end{figure}

To address this challenge, we propose to robustify the model at inference time. 
% improve the framework to make the inference algorithm more robust at inference time.
Specifically, instead of retraining the whole model on the new distribution, our inference-time defense shifts the burden to test time with our robust inference algorithm without updating the model. Prior work~\cite{mao2021adversarial, shi2020online, wang2020tent} is limited to a single constraint at inference time and hence may not provide the model with enough information to dynamically adjust to the unique and potentially novel characteristics of the corruption in the testing image. We therefore ask the natural question: \emph{Can we further improve the robustness through increasing the number of constraints?}

We start with theoretical analysis and prove that applying more constraints at inference time strictly improves the model's robustness. The next question is then: \emph{How to efficiently apply multiple constraints at inference time?} 
One approach is to directly apply multiple feature invariance constraints to the defense. While this defense is effective, we find the resulting representation can be limited by the invariance property, therefore harming robust accuracy. For example, after resizing, the representations of the segmentation models are not the same, and it is unclear which part should be invariant.

\begin{figure*}[t]
\centering
\includegraphics[width=\textwidth]{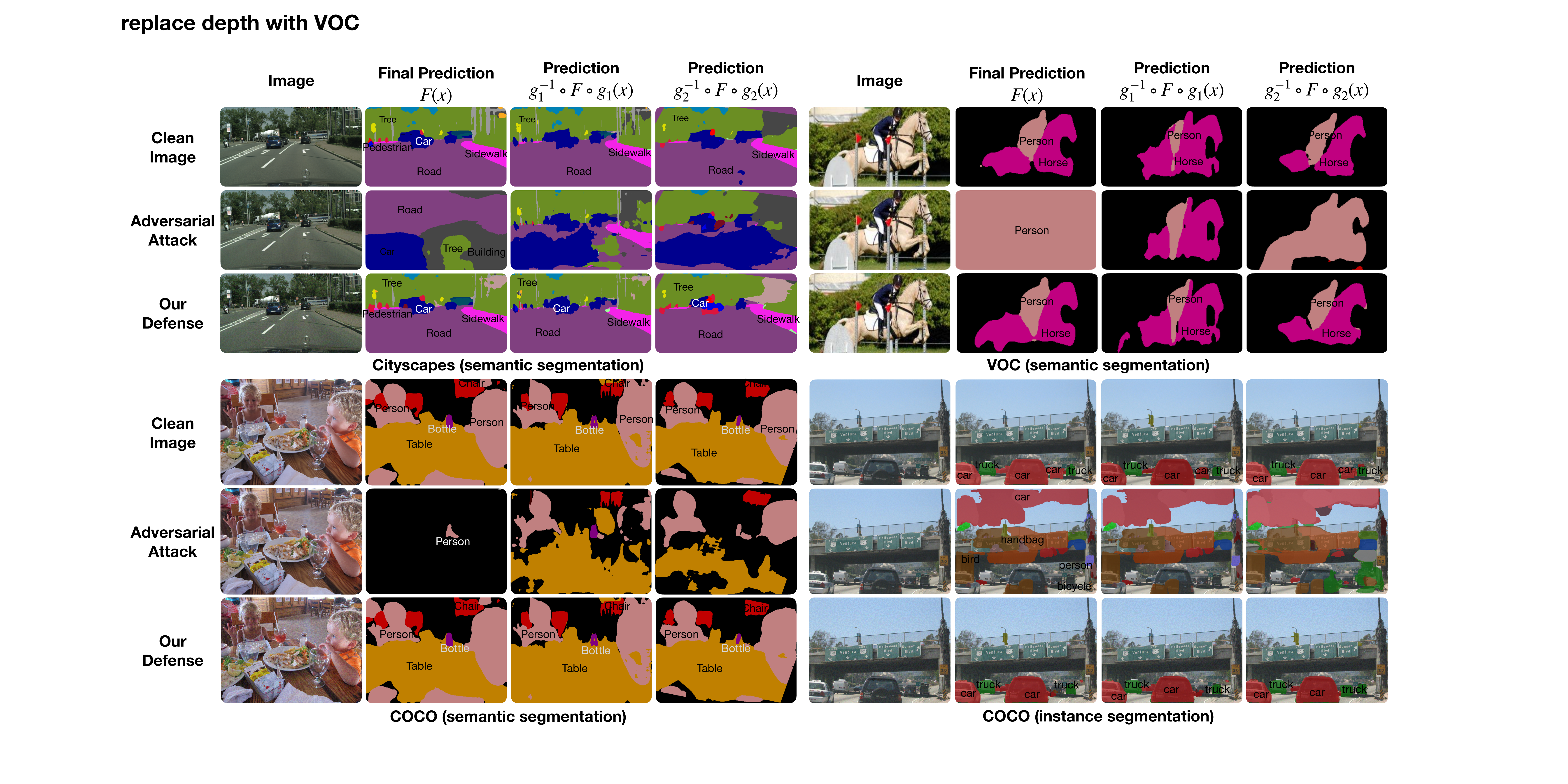}
%   \includegraphics[width=0.8\textwidth]{figures/result_lambda_matrices.pdf}
% \vspace{-5mm}
  \caption{\small{Random examples showing equivariance on clean images and non-equivariance on attacked images in Cityscapes, PASCAL VOC, and COCO. The representation is equivariant when the predicted images (2nd column) and the reversed prediction of transformed images (3rd, 4th column) are the same.  By restoring equivariance, our method corrects the prediction.}}
  \label{fig:teaser}
\end{figure*}

Our further study with empirical results suggest that a better approach is to use dense equivariance constraints. 
% In this paper, we propose to improve the robustness of models through dense equivariance constraints.
Our main hypothesis is that visual representations must be equivariant under spatial transformations, which is a dense property that should hold for all natural images (equivariance consistency in Figure~\ref{fig:equi}). This property holds when the test data are from the same distribution that the model has been trained on. However, once there has been an adversarial corruption, the equivariance is often broken  (Figure~\ref{fig:teaser}). Therefore our key insight is that we can repair the model's prediction on corrupted data by restoring the equivariance. %, and detecting uncertain data when equivariance is not respected. 

Empirical experiments, theoretical analysis, and visualizations highlight that equivariance significantly improves model robustness over other methods~\cite{mao2021adversarial, shi2020online, wang2020tent}. On four large datasets  (ImageNet~\cite{imagenet_cvpr09}, Cityscapes~\cite{cordts2016cityscapes}, PASCAL-VOC~\cite{everingham2010pascal}, and MS-COCO~\cite{COCO}), our approach improves adversarial robust accuracy by up to 15 points. Our study shows that equivariance can efficiently improve robustness  by increasing the number of constraints (Figure~\ref{fig:multiconstraints}). Even under two adaptive adversarial attacks where the attacker knows our defense~\cite{athalye2018obfuscated, mao2021adversarial}, adding our method improves robustness. In addition, since equivariance is an intrinsic property of visual models, we do not need to train a separate model to predict equivariance~\cite{shi2020online, mao2021adversarial}.  Our code is available at \url{https://github.com/cvlab-columbia/Equi4Rob}.

% We will release the model and code.

% Moreover, using equivariance improves detection efficiency, measured by AUROC,  by up to 25 points. 
% using a single constraint

\section{Related Work}

\textbf{Equivariance.} Equivariance benefits a number of visual tasks~\cite{dieleman2016exploiting, cohen2016steerable, gupta2021rotation, zhang2019making, chaman2021truly, chaman2021truly2}. \citet{cohen2016group} proposed the first group-convolutional operation that produces equivariant features to symmetry-group. However, it can only be equivariant on a discrete subset of transformation~\cite{sosnovik2019scale}. Steerable equivariance achieves continuous equivariant transformation~\cite{cohen2016steerable, weiler2018learning} on the defined set of basis, but they cannot be applied to arbitrary convolution filters due to the requirement of an equivariant basis. 
Besides architecture design~\cite{weiler2019general}, adding regularization~\cite{barnard1991invariance_trainforequi} can improve equivariance in the network. \cite{kamath2021can} shows that training equivariance at training time decreases adversarial robustness. Our method sidesteps this issue by promoting equivariance for attacked images at test time, improving equivariance when it is most needed.

% Our hypothesis is that the model is burdened at training time with both equivariance and the task of interest, where we instead propose to overcome this by fixing equivariance at inference time.

\textbf{Adversarial Attack and Defense.} Adversarial attacks~\cite{szegedy2013intriguing, madry2017towards, cisse2017houdini, mim, CW, AA, arnab2018robustness} are perturbations optimized to change the prediction of deep networks. Adversarial training~\cite{madry2017towards, rice2020overfitting, unlabeled} and its variants~\cite{mao2019metric, mao2022understanding, zhang2019theoretically} are the standard way to defend adversarial examples. The matching algorithm to produce features invariant to adversarial perturbations has been shown to produce robust models~\cite{mahajan2021domain, zhang2019theoretically}. However, training time defense can only be robust to the attack that it has been trained on. Multitask learning~\cite{mao2020multitask, zamir2020robust} and regularization~\cite{cisse2017parseval} can improve adversarial robustness. However, they did not consider the spatial equivariance in their task. Recently, inference time defense using contrastive invariance~\cite{mao2021adversarial} and rotation~\cite{shi2020online} has been shown to improve adversarial robustness without retraining the model on unforeseen attacks. However, they only apply a single constraint, which may not provide enough information.

\textbf{Test Test Adaptation.} \citet{berthelot2019mixmatch, pastore2021closer} perform test-time training on the entire test set for many iterations, our method only assumes seeing one example at a time and performs test-time adaptation on a single image. \citet{tsai2023self} adapts the model with convolutional prompt, but only works for a large batchsize. Test-time adaption is also useful in language domain~\cite{mcdermott2023robustifying}. By leveraging equivariance, we can efficiently incorporate dense constraints into our framework, which can be orders of magnitude more effective than adding constraints individually~\citep{sun2020test, lawhon2022using}.

% 
% Different from existing approaches to learning a neural network model that is equivariant on the training set, our work updates the input so that they can produce equivariant features on the given network. 

% Models with equivariance have shown excellent performance and data efficiency on vision and medical imaging problems that exhibit symmetries. Existing work obtains equivariance by designing model architecture or . The popular convolution operation respects the translation equivariance.  Designing equivariance model architecture is challenging, group-equivariant convolution has been proposed to produce a scale-equivariant network,  Existing work can only be scale equivariance~\cite{sosnovik2019scale} or translation equivariant in discrete space, but not achieving all in continuous space~\cite{tai2019equivariant}. 

% \textbf{Consistency and Robustness.} Cross task consistency~\cite{zamir2020robust} improves models' robustness. TRADES~\cite{zhang2019theoretically} improves robustness by matching the representation of clean examples and the corresponding adversarial examples to be consistent. . \cite{mao2021adversarial} restore robustness by recovering test time feature consistency with contrastive loss that captures invariance. However, invariance does not generalize to complex prediction tasks such as segmentation. In this paper, we use equivariance as a consistency to improve robustness. 

\section{Method}

In this section, we first introduce equivariance for visual representation, present algorithms to improve adversarial robustness using equivariance, and then provide theoretical insight into why the multiple constraints can lead to such improvement.  % both anomaly detection and adversarial robustness.

% We first introduce equivariance for visual representation. We then show anomaly detection using equivariance measurement, and present the algorithm to improve robustness on adversarial examples. We lastly provide theoretical insight on why equivariance improves both anomaly detection and adversarial robustness.

\subsection{Equivariance in Vision Representation}
Let $\x$ be an input image. A neural network produces a representation $\h=F_{\theta}(\x) $ for the input image. Assume there is a transformation $g$ for the input image. A neural network is equivariant only when:
% \begin{equation}
%      F_{\theta}(\x) = g^{-1} \cdot F_{\theta}(g \cdot \x),
% \end{equation}
\begin{equation}
     F_{\theta}(\x) = g^{-1} \circ F_{\theta} \circ g (\x),
\end{equation}
where $g^{-1}(\cdot)$ denotes the inverse transformation for $g(\cdot)$, and $\circ$ denotes function composition. 
Equivariant representations will change symmetrically as the input transformation. This means applying the transformation to the input image and undoing the transformation in the representation space, should result in the same representation as fed in the original image. Equivariance provides a meta property that can be applied to dense feature maps, and generalized to most existing vision tasks~\cite{gupta2021rotation, laptev2016ti, marcos2016learning}.

In contrast, invariance is defined as $F_{\theta}(\x) =   F_{\theta} \circ g (\x)$, which requires the model to produce the same representation after different transformations, such as texture augmentation~\cite{stylized-imgnet} and color jittering~\cite{mao2021adversarial, chen2020simple}. Without performing transformation in the same way as the input, invariance removes all the information related to the transformation, which can hurt the final task if the transformation is crucial to the final task~\cite{lee2021improving}.
On the contrary, equivariant models maintain the covariance of the transformations~\cite{gupta2021rotation, laptev2016ti, marcos2016learning}.

\textbf{Transformation for Equivariance.} We use spatial transformation, such as flip, resizing, and rotation, in our experiments.
Assume we apply $k$ different transformations $g_i$ where $i=1,..., k$. We denote the cosine similarity as $cos(\cdot)$. Equivariance across all transformations means the following term is large:

\begin{equation}\label{eq:equiloss}
    \mathcal{L}_{equi} = \sum\nolimits_{i=1}^k cos(g_i^{-1} \circ F_{\theta} \circ g_i(\x), F_{\theta}(\x))
\end{equation}

\begin{algorithm}[t]
\caption{Equivariance Defense}
\label{algorithm: defense}
\begin{algorithmic}[1]
\STATE {\bfseries Input:} Potentially attacked image $\x$, step size $\eta$, number of iterations $T$, deep network $F$, reverse attack bound $\epsilon_v$, and equivariance loss function $\mathcal{L}_{equi}$.
\STATE {\bfseries Output:}  Prediction $\hat{y}$
% \STATE{Randomly initialize network $f_\theta$, or initialize network with pre-trained configuration}
\STATE{\bfseries Inference: $\x'\leftarrow \x$}
\FOR{$t=1,...,T$}
\STATE{$\x'\leftarrow \x' + \eta \cdot \text{sign}( \text{Normalize}(\nabla_{\x'} \mathcal{L}_{equi}(\x')) + \mathcal{N}(0, \frac{T-1-t}{T}))$}
\STATE{$\x'\leftarrow \Pi_{\mathbb(\x,\epsilon_v)} \x' $, which projects the image back into the bounded region.} %\bbB throws 'undefined control sequence'
\ENDFOR 
\STATE{Predict the final output by $\hat{y}=F(\x')$}

% \STATE{$\theta\leftarrow \theta-\eta_2 \sum_{i=1}^m \nabla_\theta[\cL(f_\theta(\x_i),\y_i)+\cL(f_\theta(\x_i),f_\theta(\x_i'))/\lambda]/m$}

\end{algorithmic}
\end{algorithm}

\subsection{Equivariance for Adversarial Robustness}
Let $\y$ be the ground-truth category labels for $\x$. Let the network that uses the feature $\h$ for final task prediction to be $C_{\theta'}$. For prediction, neural networks learn to predict the category $\hat{\y}=C_{\theta'} \circ F_{\theta}(\x) $
by minimizing the loss $L(\hat{\y}, \y)$ between the predictions and the ground truth. For example, for semantic segmentation, $L$ is cross-entropy for each pixel output. We define the loss for the final task as follows:
\begin{equation}\label{eq:loss_task}
    \mathcal{L}_t(\x, \y) = L \left(C_{\theta'} \circ F_{\theta}(\x), \y\right),
\end{equation}

\textbf{Adversarial Attack.} To fool the model's prediction, the adversarial attack finds additive perturbations $\boldsymbol{\delta}$ to the image such that the loss of the task (Equation~\ref{eq:loss_task}) is maximized. %
\begin{equation}
    \x_{a} = \argmax_{\x_{a}} \mathcal{L}_t(\x_{a}, \y), \quad \text{s.t.} \quad ||\x_{a}-\x||_q \leq \epsilon,
    \label{eq:advattack}
\end{equation}
where the perturbation vector $\boldsymbol{\delta}=\x_{a}-\x$ has a $q$ norm bound that is smaller than $\epsilon$, keeping the perturbation invisible to humans. 

\textbf{Equivariance Recalibration Defense.} Given an input image, we can calculate the equivariant loss $\mathcal{L}_{equi}$. As shown in Figure~\ref{fig:teaser}, the representations are non-equivariant when the input $\x_a$ is adversarially perturbed, i.e., the term $\mathcal{L}_{equi}$ is low. We will find an intervention to recalibrate the input image $\x_a$ such that we can improve the feature equivariance of the image. To do this, we optimize a vector $\r$ by maximizing the equivariance objective:
\begin{equation} \label{eq:reverse}
    \r = \argmax_{\r} \mathcal{L}_{equi}(\x_{a} + \r), \quad \text{s.t.} \quad ||\r||_q \leq \epsilon_v,
\end{equation}
where $\epsilon_v$ defines the bound of our reverse attack $\r$. The additive intervention $\r$ will modify the adversarial image $\x_{a}$ such that it restores the equivariance in feature space.

We optimize the above objective via projected gradient descent to repair the input. To avoid the optimization converging to a local optimal, we first perform SGLD~\cite{sgld} to get a good Bayesian posterior distribution of the solution~\cite{wang2019bidirectional}. To avoid sampling from the posterior distribution of SGLD and improve the inference speed, we then use maximum a posterior (MAP) estimation to find a single solution. Empirically, we add Gaussian noise to the gradient when optimizing and linearly anneal the noise level to zero. We show the optimization procedure in Algorithm~\ref{algorithm: defense}. We use the same optimizer for the invariance objective and compare.

In contrast to \citet{mao2021adversarial, shi2020online}, we do not need to pre-train another network for the self-supervision task offline. In addition, equivariance in the feature space provides a dense constraint because, by projecting the transformation back to the original space, we can match each element in the feature space. Image-level self-supervision tasks, such as contrastive loss and rotation prediction, do not have this dense supervision advantage.

% , because it cannot match the features of spatial transformation without first pooling them to the same space.

\textbf{Adaptive Attack I.} We now analyze our methods' robustness when the attacker knows our defense strategy and takes our defense into consideration. Following the defense-aware attack setup in~\cite{mao2021adversarial}, the adaptive attacker can maximize the following equation:
\begin{equation}
\label{eq:ada_attack}
    \mathcal{L}_l(\x_{a}, \y, \lambda_s)=\mathcal{L}_t(\x_{a}, \y) + \lambda_e \mathcal{L}_{equi}(\x_{a}).
\end{equation}
where the first term fools the final task, and the second term optimizes for equivariance. A larger $\lambda_e$ allows the adversarial budget to focus more on respecting the feature equivariance, which reduces the defense capability of our defense. However, with a fixed adversarial budget, increasing $\lambda_e$ also reduces the attack efficiency for the final task. Our defense creates a lose-lose situation for the attacker. If they consider our defense, they hurt the attack efficiency for the final task. If they ignore our defense, our defense will fix the attack.

\textbf{Adaptive Attack II.} The above adaptive attack avoids the unstable gradient from the iterative optimization with a Lagrangian regularization term. Another way to bypass such defense is through BPDA~\cite{athalye2018obfuscated}. Specifically, the equivariance recalibration process formulated in Eq.~\ref{eq:reverse} can be treated as a preprocessor $h(\cdot)$ that is employed at test time, where $h(\x_a) = \x_a + \r$. Given a pre-trained classifier $f(\cdot)$, this method can be formulated as $f(h(x))$. The proposed process $h(\cdot)$ may cause exploding or vanishing gradients. According to ~\cite{athalye2018obfuscated, croce2022evaluating}, we can use BPDA to approximate $h(\cdot)$, where an identity function is used for the backward pass of the restored images. While this method may make the backward gradient inaccurate, it avoids differentiation through the inner optimization procedure, which often leads to vanished or exploded gradients.

\begin{figure}
\centering
 % \vspace{} 
  \includegraphics[width=0.9\linewidth]{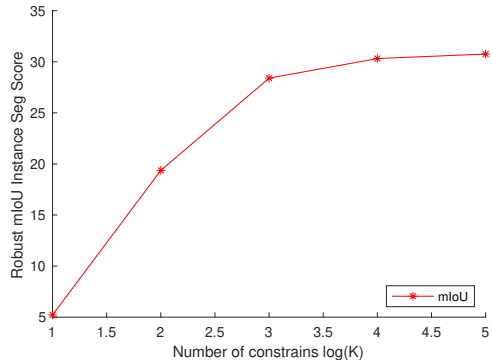}
  \caption{Adversarial robustness under an increased number of constraints through equivariance at inference time.} \vspace{-5mm} \label{fig:multiconstraints}
\end{figure}

\subsection{Theoretical Results for Adversarial Robustness with Multiple Constraints}
One major advantage of equivariance is that it allows dense constraints through the inverse transformation.
We show theoretical insights for why using a dense intrinsic constraint rather than a single intrinsic constraint.
Existing methods restore the input image to respect a single self-supervision label $y^{(s_1)}$. With a dense intrinsic constraint, the defense model can predict with a set of fine-grained self-supervision signals. $y^{(s_i)}$, where $i=1, 2, ..., K$. In our case, each $y_a^{(s_i)}$ is the predicted self-supervision value under adversarial attack, and each $y^{(s_i)}$ is the predicted self-supervision value in our feature map after equivariance transformation.
Following~\cite{mao2021adversarial}, we propose the following lemma:
\begin{lemma}
The standard classifier under adversarial attack is equivalent to predicting with $ P(\Y | \X_a, y^{(s_1)}_a, y^{(s_2)}_a, ..., y^{(s_k)}_a)$, and our approach is equivalent to predicting with $P(\Y | \X_a, y^{(s_1)}, y^{(s_2)}, ..., y^{(s_k)})$.
\end{lemma}
By adjusting the input image such that it satisfies a set of denser constraints, the predicted task $Y$ uses both the information from the image and the intrinsic equivariance structure. We now show that by restoring the dense constraints in our visual representation, from an information perspective, the upper bound can be strictly improved than just restoring the structure from a single self-supervision task~\cite{mao2021adversarial, shi2020online}.

% Hao: remember to add an additional MI condition
\begin{theorem}\label{thm:boundMTL1}
Assume the classifier operates better than chance and instances in the dataset are uniformly distributed over $n$ categories. Let the prediction accuracy bounds be $P(\Y|y^{(s_1)}_a, y^{(s_2)}_a, ..., y^{(s_k)}_a, \X_a) \in [b_0, c_0]$, $P(\Y|y^{(s_1)}, \X_a) \in [b_1, c_1]$,  $P(\Y|y^{(s_1)}, y^{(s_2)}, \X_a) \in [b_2, c_2]$, ..., and $P(\Y|y^{(s_1)}, y^{(s_2)}, ..., y^{(s_k)},\X_a) \in [b_k, c_k]$. If the conditional mutual information $I(\Y;\Y^{(s_i)}| \X_a) > 0$ and $I(\Y;\Y^{(s_i)}| \X_a, \Y^{(s_j)}) > 0$ where $i \neq j$, we have $b_0 \leq b_1 \leq ... \leq b_k$ and $c_0 < c_1 < c_2 < ... < c_k$, which means our approach strictly improves the upper bound for classification accuracy.
\end{theorem}
In words, the adversarial perturbation $\X_a$ corrupts the shared information between the label $\Y$ (our target task) and the equivariance structure $\Y^{s_i}$ (self-supervised task). Theorem \ref{thm:boundMTL1} shows that by recovering information from more $\Y^{s_i}$, the task performance can be improved.

Directly increasing the number of invariance objectives is a straightforward baseline to increase the number of constraints. However, this can be limited because 1) each invariance objective only adds one constraint, which is less efficient, and 2) invariance cannot be directly applied to many transformations, such as resizing and rotation, due to the mismatch in fine-grained representation, where equivariance can.
In contrast to invariance, dense equivariance allows us add one constraint on each element in the feature map\footnote{A 100 by 100 feature map would provide 10000 constraints}, which can increase constraints orders of magnitude faster with a more diverse set of transformations, providing an efficient way to apply multiple constraints. By subsampling different number of constraints from equivariance, Figure~\ref{fig:multiconstraints}  validates the trend of improving robustness as the number of constraints increases.

The adaptive attack needs to respect the information in $Y^{s_i}$, which itself limits the ability of the attacker, as the attacker performs a multitask optimization which is harder~\cite{mao2020multitask}. The adaptive attacker predicts the task conditioned on the right set of self-supervision label $\Y^{s_i}$, which fulfills our Theorem~\ref{thm:boundMTL1} and improves robustness.

\section{Experiments}
Our experiments evaluate the adversarial robustness on four datasets: ImageNet~\cite{imagenet_cvpr09}, Cityscapes~\cite{cordts2016cityscapes}, PASCAL-VOC~\cite{everingham2010pascal}, and MS-COCO~\cite{COCO}. We use up to 6 different strong attacks, including Houdini, adaptive attack, and BPDA to evaluate the robustness. We first show that our equivariance-based defense improves the robustness of the state-of-the-art adversarially trained robust models. We then show that even on the standard models without defense training, adding test-time equivariance can improve their robustness.

\begin{table*}[t]
\centering
% \vspace{-3mm}
\caption{{Classification accuracy on ImageNet and segmentation mIoU on Cityscapes dataset on adversarially trained models with $\epsilon=4/255$. Using equivariance improves robustness more than other methods.
}} % attack, step size =1, eps=4.
% \small
    \begin{tabular}{l|c|c|c|c|c|c}
        \toprule
         & \multicolumn{6}{c}{ImageNet; Adversarially Pretrained Model~\cite{wong2020fast};  Classification Accuracy} \\
          Evaluation Method    & Vanilla & Random & Rotation & Contrastive & Invariance & Equivariance (Ours)   \\
         \midrule 
         Clean & \textbf{51.5} & 49.4 & 49.5 & 49.2 & 48.8 & 49.3\\
        %  \midrule
         PGD & 26.5 & 28.0 & 28.2 & 29.3 & 28.6 & \textbf{32.2}\\
        %  PGD-200 & 1.56 & 1.68 & 15.64 & 9.56 & 15.86 & \textbf{30.99}\\
         CW & 26.6 & 28.3 & 28.6 & 29.8 & 32.2 & \textbf{32.2}\\
         AA & 26.5 & 28.0 & 28.2 & 29.3 & 28.6 & \textbf{32.2} \\
         BPDA & 26.5 & 28.0 & 27.9 & 28.8 & 28.9 & \textbf{30.4} \\  % 27.8 for vanilla
         \midrule
         \toprule
         & \multicolumn{6}{c}{Cityscape; Adversarially Trained DRN-22-d; Segmentation MIoU} \\
          Evaluation Method    & Vanilla & Random & Rotation & Contrastive & Invariance & Equivariance (Ours)   \\
         \midrule 
         Clean & \textbf{53.23} & 52.96 & 51.72 & 53.00 & 49.04 & 48.74 \\
         IFGSM (seg) & 33.06 & 33.21 & 33.47 & 33.59 & 32.36 & \textbf{34.04}\\
         PGD & 26.61 & 27.04 & 27.68 & 28.14 & 27.74 & \textbf{29.65} \\
         MIM & 26.59 & 27.06 & 27.72 & 28.24 & 27.76 & \textbf{29.56}\\
         Houdini & 23.47 & 24.07 & 25.56 & 26.61 & 26.97 & \textbf{29.80} \\
         AA & 26.61 & 27.04 & 27.68 & 28.14 & 27.74 & \textbf{29.63}\\
         BPDA & 26.61 & 27.00  & 26.37 & 28.50 & 23.31 & \textbf{29.83}  \\
         \bottomrule
    \end{tabular}
\vspace{-3mm}
\label{tab:adv-city}
\end{table*} % 20 steps or reversal , stepsize=2, 1.5 mul.

% \subsection{Robustness against Adversarial attacks}

\subsection{Dataset and Tasks}
\emph{ImageNet} \cite{imagenet_cvpr09} contains 1000 categories. Due to its large size, we randomly sample 2\% of data for evaluation. \emph{Cityscapes}~\cite{cordts2016cityscapes} is a urban driving scene dataset. We study the {semantic segmentation} task. Following~\cite{mao2020multitask}, we resize the image to $680 \times 340$ for fast inference. We use pretrained dilated residual network (DRN) for segmentation.
\emph{PASCAL-VOC}~\cite{everingham2010pascal} is a dataset for {semantic segmentation} task. We resize images to $480 \times 480$. We use the pre-trained DeepLabV3+ model.
\emph{MS-COCO}~\cite{COCO} is a large-scale image dataset of common objects that supports {semantic segmentation} and {instance segmentation} task. For semantic segmentation, we resize the images to $400 \times 400$. We use pretrained DeeplabV3 and MaskRCNN for semantic segmentation and instance segmentation, respectively.

\subsection{Attack Methods} 
\emph{IFGSM (seg)}~\cite{arnab2018robustness} was used to evaluate the robustness of segmentation models with multiple steps of the fast gradient sign method.
\emph{PGD}~\cite{madry2017towards} is a standard iterative-based adversarial attack, which performs gradient ascent and projects the attack vector inside the defined $p$ norm ball.
\emph{MIM}~\cite{mim} adds a momentum term to the gradient ascent of PGD attack, which is a stronger attack that can get out of local optima.
\emph{Houdini}~\cite{cisse2017houdini} is the state-of-the-art adversarial attack for decreasing the mIoU score of semantic segmentation. It proposes a surrogate objective function that can be optimized on the mIoU score directly.
\emph{Adaptive Attack (AA)}~\cite{mao2021adversarial} is the standard defense-aware attack for inference time defense method, where the adaptive attack knows our defense algorithm, and optimizes the attack vector to respect equivariance while fooling the final task. Since the attack already respects and adapts to equivariance, our defense has less space to improve by further optimizing for equivariance.
\emph{BPDA}~\cite{athalye2018obfuscated, croce2022evaluating} is an adaptive attack for input purification. In our case, we forward the adapted images in the forward pass and straight-through the gradient from our adapted image to the input image.

\subsection{Baselines} 
We compare our method with the vanilla feed-forward inference and four existing inference-time defense methods. 
\emph{Random} defense~\cite{random} defends adversarial attack by adding random noise to the input, which is used as a baseline in \cite{mao2021adversarial}. 
\emph{Rotation} defense~\cite{shi2020online} purifies the adversarial examples by restoring the performance of the rotation task at inference time, which can recover the image information that relates to rotation. However, the information related to rotation may be misleading due to the illusion issue~\cite{illusion}, which limits its power for complex tasks.
\emph{Contrastive} defense~\cite{mao2021adversarial} restores the intrinsic structure of the image using SimCLR~\cite{chen2020simple} objective at inference time, which achieves state-of-the-art adversarial robustness on image recognition tasks. Contrastive learning requires images to be object-centric, which may not be true on the segmentation and detection dataset where multiple objects appear in the same image.
\emph{Invariance} defense follows the same setup as our equivariance experiment but replaces the equivariance loss with the invariance loss. To obtain several constrants from invariance, we use the same diversified set of transformations as the equivariance setup. We propose this baseline to study the importance of using equivariance to apply multiple constraints.

\subsection{Implementation details}\label{sec:details}
We choose the number of transformations to be $K=8$, which empirically can be fit into a 2080Ti GPU with batch size 1. To increase the constraints obtained from equivariance, we empirically use a diversified set of transformations, which includes four resizing transformations ranging from 0.3 to 2 times of size change; one color jittering transformation; one horizontal flip transformation; and two rotation transformations between -15 to 15 degrees. For transformations that cause part of the original image not in the view, we only consider the overlapped region when calculating the loss. Ablation study for the effect of each transformation is shown in Section~\ref{subsec:analysis}. We use steps $T=20$ for all our defense tasks. Since after the spatial transformations, the invariance objective cannot be performed in the dense feature space due to the position mismatch, we apply an average pooling for all the features and then compute the invariance loss.

% except for the depth where we use $T=50$ steps.

\begin{table*}[t]
\centering
% \scriptsize
% \scriptsize
\caption{{Semantic segmentation mIoU on Cityscapes, PASCAL VOC, and MSCOCO dataset. All models are not adversarially trained. Under different types of attack bounded by $L_{\infty}=4/255$, our method consistently outperforms other defense methods. 
}} % attack, step size =1, eps=4.
% \vspace{-3mm}
    \begin{tabular}{l|c|c|c|c|c|c}
        
         \toprule
         & \multicolumn{6}{c}{Cityscape; Pretrained DRN-22-d Model; Segmentation MIoU} \\
          Evaluation Method   & Vanilla & Random & Rotation & Contrastive & Invariance & Equivariance (Ours)   \\
         \midrule 
         Clean & 58.29 & 52.38 & 34.30 & 37.22 & 33.84 & 37.95\\
        %  \midrule
         PGD & 1.31 & 1.47 & 13.20 & 8.44 & 14.49 & \textbf{30.76} \\
        %  PGD-200 & 1.56 & 1.68 & 15.64 & 9.56 & 15.86 & \textbf{30.99}\\
         MIM & 1.40 & 1.49  & 13.80 & 8.13 & 14.57 & \textbf{30.10}\\
        %  MIM-200 & 1.60  & 1.64  & 16.17 & 9.86 & 14.50 & \textbf{30.68}\\
         Houdini & 0.00 & 0.21 & 16.31 & 10.12 & 14.16 & \textbf{30.52}\\
         AA & 1.31 & 1.47 & 13.20 & 8.44 & 14.49 & \textbf{30.28} \\
         % BPDA & 1.31 & 1.32 & 4.62 & 3.53 & 8.93 & \textbf{14.80} \\
         BPDA & 1.31 & 1.47 & 6.20 & 4.78 & 6.82 & \textbf{11.64} \\ % BPDA step 10, reverse 20, so together 200 steps of optimization, epsilon = 1.5v
         \bottomrule
         \toprule
         & \multicolumn{6}{c}{ PASCAL VOC dataset; Pretrained DeepLabV3; Segmentation MIoU} \\
          Evaluation Method    & Vanilla & Random & Rotation & Contrastive & Invariance & Equivariance (Ours)   \\
         \midrule 
         Clean & 69.52 & 68.96 & 28.63 & 66.92 & 63.64 & 56.58 \\ % 50 steps
         PGD & 6.46 & 6.52 & 6.91 & 18.72 & 39.07 & \textbf{43.51} \\ % 50 steps
         MIM & 5.63 & 5.74 & 6.35 & 18.25 & 37.43 & \textbf{41.56} \\ % 50 steps
         Houdini & 0.02 & 0.08 & 6.14 & 19.11 & 31.30 & \textbf{52.26} \\ % 50 steps
         % AA & - &-&- &- &- & \textbf{43.85} \\ % 50 steps
         BPDA & 6.46 & 6.46 &8.23 &5.45 &15.15 & \textbf{25.68} \\ % 50 steps
         \bottomrule
         \toprule
         & \multicolumn{6}{c}{MSCOCO dataset; Pretrained DeeplabV3-resnet50; Segmentation MIoU} \\
          Evaluation Method    & Vanilla & Random & Rotation & Contrastive & Invariance & Equivariance (Ours)   \\
         \midrule 
         Clean & 63.02 & 62.97 & 60.92 & 57.28 & 43.07 & 44.71\\
         PGD & 2.62  & 2.65 & 5.79 & 14.75 & 23.92 & \textbf{24.51}\\
         MIM & 2.71 & 2.52 & 5.66 & 13.61 & 20.53 &\textbf{21.30}\\
         Houdini &  0.05 & 0.10 & 4.78 & 22.69 & 36.94 & \textbf{37.33}\\
         % AA & - &-&- &- &- & \textbf{23.91}\\
         BPDA & 2.62 & 2.63 & 1.15 & 2.35 & 17.13 & \textbf{18.69}\\
         \bottomrule
         \toprule
         & \multicolumn{6}{c}{MaskRCNN; Instance Segmentation maskAP} \\
          Evaluation Method    & Vanilla & Random & Rotation & Contrastive & Invariance & Equivariance (Ours)   \\
         \midrule 
         Clean & 34.5 & 33.6 & 31.2 & 29.7 & 14.3 & 23.4\\
         PGD & 0.0 & 1.6 & 2.6 & 8.9 & 12.9 & \textbf{21.3}\\
         MIM & 0.0 & 1.6 & 2.7 & 9.1 & 13.2 & \textbf{21.2}\\
        %  Houdini \\
         % AA & - &-&- &- &- &\textbf{23.5}\\
         BPDA & 0.0 & 0.0 & 0.3 & 1.7 & 8.7 & \textbf{9.9} \\
         \bottomrule
    \end{tabular}
% \vspace{-7mm}
\vspace{-4mm}
\label{tab:city-seg}
\end{table*} %  stepsize=2, 1.5 mul.

% We also show that our approach complements adversarial training

\begin{table*}[t]
\centering
% \scriptsize
\caption{\small{Segmentation mIoU with targeted attack on Cityscapes. We use the DRN-22-d backbone. Restoring the equivariance moves the predicted segmentation map to the groundtruth.
}}
% \vspace{-3mm}
    \begin{tabular}{l|c|c|c||c|c|c}
         \toprule
         % & \multicolumn{6}{c}{DRN-22-d} \\
         & \multicolumn{3}{c}{mIoU to Attack Target $\downarrow$} & \multicolumn{3}{c}{mIoU to Groundtruth $\uparrow$}\\
          Evaluation Method   & Vanilla  & Invariance & Equivariance (Ours) & Vanilla  & Invariance & Equivariance (Ours)   \\
         \midrule 
         PGD & 68.03 & 12.92 & 14.96 & 10.08 & 25.10 & \textbf{30.01}\\ % steps=
         MIM & 71.49 & 13.78 & 12.63 & 9.78 & 23.11 & \textbf{28.51}\\
         Houdini & 54.49 & 12.83 & 16.80 & 17.17 & 24.56 & \textbf{30.26} \\
         BPDA & 68.03 & 25.64 & 25.82 & 10.08 & 17.73 & \textbf{20.14}\\
         \bottomrule
    \end{tabular}
\vspace{-5mm}
\label{tab:targeted}
\end{table*}

\subsection{Results on Adversarial Trained Models}
% White box attack is the strongest adversarial attack for deep networks, where the attacker knows everything about the model. 

% We conduct a white-box attack on three semantic segmentation datasets, one depth prediction dataset, and one instance segmentation dataset.

Adversarial training is the standard way to defend against adversarial examples. We first validate whether our proposed approach can further improve the robustness of adversarially trained models. For ImageNet, we use the adversarial pretrained model with $\epsilon=4/255$ from \cite{wong2020fast}. We set the defense vector bound to be $\epsilon_v = 2\epsilon$. With the state-of-the-art contrastive learning method~\cite{mao2021adversarial}, we improve robustness accuracy by 3 points to the Vanilla defended model. Adaptive attack (AA) poses a lose-lose situation and does not further decrease the robustness accuracy, which is consistent with the observation of~\cite{mao2021adversarial}. With the strongest adaptive attack BPDA~\cite{croce2022evaluating}, it drops 0.5 points. \footnote{Recent work~\cite{croce2022evaluating} uses a batch size of 50 for contrastive loss, which is a weaker defense due to the small batch size. Here, we use the original batch size of 400 setup as \cite{mao2021adversarial}, which provides a stronger defense due to the large batch size, where we see robust accuracy improved than Vanilla.} Using the equivariance objective, under both standard attack and the adaptive attack BPDA, it improves robustness more than the other methods. Even though BPDA decreases equivariance defense by 1.8 points, equivariance still improves robustness by 3.9 points than not using it.
 
 On Cityscapes, we downsample the image from 2048 $\times$ 1024 to 680 $\times$ 340 to reduce computation, which follows the setup of \cite{mao2020multitask}. We adversarially train a segmentation model and evaluate it in Table~\ref{tab:adv-city}, which is measured with mean Intersection over Union (mIoU) for semantic segmentation. We set the defense vector bound to be $\epsilon_v = 2.5\epsilon$. For the standard attack, Houdini reduces the robustness accuracy the most, where using equivariance constraints at test time can recover 6 points of performance. Using the adaptive attack~\cite{mao2021adversarial}, the robust accuracy of equivariance only drops by 0.2 points. Using the BPDA adaptive attack, the robustness of the invariance-based method drops 4 points, which suggests that invariance relies mostly on obfuscated gradients and it is not an effective constraint to maintain at inference time for segmentation. In contrast, BPDA \emph{cannot} undermine the equivariance-based model's robustness. On the adversarially trained model, equivariance consistently outperforms all other test-time defenses, which demonstrates that equivariance is a better intrinsic structure to respect during inference time.

\begin{SCfigure*}[1][t]
    \centering
    \includegraphics[width=.6\textwidth]{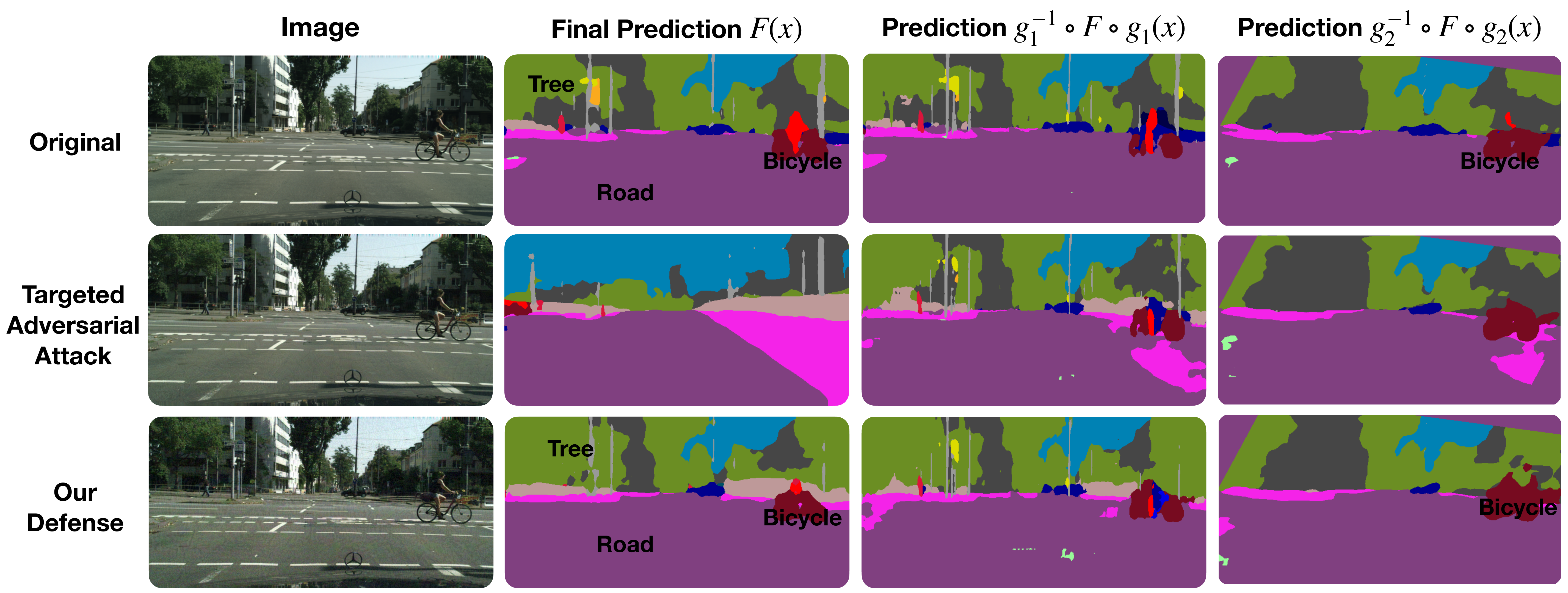}
    \caption{\small{Our method improves robustness under targeted adversarial attacks (Random Sample). By adding targeted adversarial attacks, the model fails to predict the bicycle on the road and instead predicts a sidewalk. In the middle row, the attacked model's representation produces different segmentation maps under different transformations, suggesting that the model is no longer equivariant. By restoring the equivariance, we correct the model prediction.}}
    \label{fig:targeted}
\end{SCfigure*}

\subsection{Results on Non-Adversarial Trained Models}

We have shown that equivariance improves the robustness of adversarially trained models. However, most pretrained models are not adversarially defended. We thus study whether our method can also improve standard models' robustness.

\noindent\textbf{CityScapes Semantic Segmentation.} In Table \ref{tab:city-seg}, we first conduct five types of attacks for the DRN-22-d segmentation model. We use 20 steps of defense, i.e., $K=20$, and use a step size of $\eta = 2\epsilon_v$, and set the defense vector bound to be $\epsilon_v = 1.5\epsilon$. While the strongest Houdini attack can reduce the mIoU score to 0, our defense can restore the mIoU score by over \textbf{30} points. For the adaptive attack, we search the optimal $\lambda_e$ that reduces the robust performance the most and find  $\lambda_e=1000$ produces the most effective attack, which still cannot bypass our defense. For baselines, we find $\lambda_e=0$ produces the most effective attack. We find for standard backbones that are not adversarially trained, BPDA is the most effective attack, we thus only evaluate on BPDA on the following datasets. We run 10 steps of BPDA with 20 steps of reversal in the inner loop, which is a totally of 200 backward steps. Under the BPDA attack, equivariance-based defense is still more effective than other methods, including the invariance-based method.

\noindent\textbf{PASCAL VOC Semantic Segmentation}. We show results in Table \ref{tab:city-seg}. We use the pretrained DeepLabV3~\cite{chen2018encoder, chen2017rethinking} model. We use $K=20$ and step size $\eta = 2\epsilon_v$, and $\epsilon_v = 1.5\epsilon$. Our approach can significantly improve the robustness compared with other methods.

\noindent\textbf{MSCOCO Semantic Segmentation.} We show results in Table \ref{tab:city-seg}. We use the pretrained DeepLabV3~\cite{chen2018encoder, chen2017rethinking} model. On COCO, we use $K=2$ and step size $\eta = 2\epsilon_v$, $\epsilon_v = 1.25\epsilon$. Using equivariance outperforms other test-time defense methods.

\noindent\textbf{Instance Segmentation.} Our defense can also secure the more challenging instance segmentation model. In Table~\ref{tab:city-seg}, our method improves instance segmentation maskAP by up to 21 points, which demonstrates that our method can be applied to a large number of vision applications.

\noindent\textbf{Targeted Attack.} The above attacks are untargeted. We also analyze whether our conclusion holds under targeted attacks, where the attacker needs to fool the model to predict a specific target. In Table~\ref{tab:targeted}, the targeted attack successfully misleads the model to predict the target, and our equivariance defense corrects the prediction to be the ground truth. Equivariance improves up to 10 points on the mIoU metric. We show visualizations in Figure~\ref{fig:targeted}.

\subsection{Analysis}\label{subsec:analysis}

\noindent\textbf{Equivariance Measurement.} We calculate the equivariance value measured by Equation~\ref{eq:equiloss} for clean images, adversarial attacked images, and our defended images. We show the numerical results in Table~\ref{tab:equi_metric}. While adversarial attacks corrupt the equivariance of the image, as shown by the lowered value in the table, our method is able to restore it. Visualizations in Figure~\ref{fig:teaser} also show our method clearly restores the equivariance under attack.

\begin{table*}[t]
\centering
% \scriptsize
\scriptsize

% \vspace{-3mm}
    \begin{tabular}{l|c|c|c|c|c|c}
        
         \toprule
             & Vanilla & Random & Rotation & Contrastive & Invariance & Equivariance (Ours)   \\
         \midrule 
         Running Time (sec/sample) & 0.016	&0.016	&0.322&	0.152&	1.632	&1.653 \\
         Memory Usage (GB) & 0.391 &0.391	&3.102	&0.731&	10.049	&10.357 \\
            
         \bottomrule
    \end{tabular}
    \caption{Running time and GPU memory usage for MS COCO semantic segmentation task. We evaluate on a single A6000 GPU.} % attack, step size =1, eps=4.
% \vspace{-7mm}
\vspace{-4mm}
\label{tab:memory}
\end{table*}

\begin{table}
\scriptsize
\centering
\begin{tabular}{l|c|c|c|c}
         \toprule
         & \multicolumn{3}{c}{Dataset} \\
             & ImageNet & Cityscapes & PASAL VOC & COCO     \\
         \midrule 
         Clean Images & 0.539& 0.694 & 0.900 & 0.901\\
         Attacked Images & 0.538 & 0.448 & 0.642 & 0.774\\
         Restored Images & \textbf{0.581} & \textbf{0.713} & \textbf{0.921} & \textbf{0.914}\\
         \bottomrule
    \end{tabular}
    \caption{{Measurement for equivariance on clean images, attacked images, and our restored images. A high score indicates better equivariance. Adversarial attack corrupts the equivariance. Our method restores the equivariance back to the same level as the clean images.}}
    \vspace{-4mm}
    \label{tab:equi_metric}
\end{table}

\begin{table}
\scriptsize
\begin{tabular}{l|c|c|c|c}
         \toprule
         & \multicolumn{4}{c}{Transformations of Equivariance} \\
          Loss   & Flip & Resize & Rotation$\leq 15^{\circ} $ & Rotation$\geq 90^{\circ} $    \\
         \midrule 
         Invariance & 9.56 & 9.90 & 9.75 & \textbf{9.60} \\
         Equivariance & \textbf{20.50} & \textbf{26.00} & \textbf{17.03} & 8.61\\
         \bottomrule
    \end{tabular}
    \caption{{The impact of using different transformations on the performance of our method. We show results from a standard segmentation model on Cityscape. }}
    \label{tab:ablation-equi}
    \vspace{-5mm}
\end{table}

% 

% The most useful types of transformations are those for which the task is meant to be equivariant, but the used model is not, f

\noindent\textbf{Ablation Study for Equivariance Transformations.} In Table~\ref{tab:ablation-equi}, we study the impact of using different transformations in our equivariance defense. We find transformations, which the model should be equivariant to but, in fact, does not due to attacks, are the most effective ones in improving robustness. For example, flipping and resizing are most effective for our studied semantic segmentation. Rotation below 15 degrees helps robustness more than rotation larger than 90 degrees. Large rotation performs worse because segmentation models are not equivariant to large rotation, even on clean data, which reduces the effectiveness of our approach. In Section~\ref{sec:details}, we empirically choose the combination of transformations that produces good empirical results for our approach.

\noindent\textbf{The Trade-off between Robustness and Clean Accuracy.} In Table~\ref{tab:tradeoff}, we show that increasing the bound $\epsilon_v$ for the defense vector creates a trade-off between clean accuracy and robust accuracy. Specifically, bound $\epsilon_v=1/255$ is a sweet spot, where one can increase robustness by 0.4 without any loss of clean accuracy. Our method allows dynamically conducting trade-off between robustness and clean accuracy by controlling the additive vector's bound.

\noindent\textbf{Runtime Analysis and GPU Memory Usage.} In Table~\ref{tab:memory}, we show the running time and GPU memory usage on our studied methods. While our method leads to longer running time and larger GPU memory usage, we believe this is a necessary trade-off to achieve the best robustness. In many important applications, sacrificing accuracy or robustness for the sake of reducing running time/memory usage would be counterproductive. To mitigate this, we also propose to first detect adversarial examples, then only perform our test-time adaptation for the detected adversarial ones.

\noindent\textbf{Detecting Adversarial Samples.} A straightforward way to speed up our inference and improve the accuracy on clean samples, is to first detect adversarial samples, and only run our algorithm on the adversarial samples. Table~\ref{tab:detect} reports the time running on COCO images with a single A6000 GPU, which shows that detection is less expensive compared to our defense and can be used to reduce our computational cost. Since test-time optimization on clean examples decrease clean performance, we can also increase the clean accuracy by first detecting the adversarial examples. In Appendix~\ref{subsubsec:detect}, we show that we can increase clean performance by only performing test-time optimization on the detected adversaries.

\begin{table}
\scriptsize
\begin{tabular}{l|c|c|c|c|c|c|c}
         \toprule
         & \multicolumn{7}{c}{Equivariance defense vector bound $\epsilon_v = i/255$} \\
          Accuracy   & i=0 & i=1 & i=2 & i=4 & i=6 & i=8  & i=10  \\
         \midrule 
         Clean & 53.23 & \textbf{53.24} & 53.09 & 52.38  & 50.62 & 48.84 & 48.74\\
         Robustness & 26.61 & 27.03 & 27.57 & 28.53 & 29.22 & 29.57 & \textbf{29.83}\\
         \bottomrule
    \end{tabular}
    \caption{{Trading-off Robustness vs. Clean Accuracy on Cityscape using our equivariance method under BPDA attack. If clean performance is important, we can simply decrease the defense vector bound to increase the clean accuracy.}}
    % \vspace{-4mm}
    \label{tab:tradeoff}
\end{table}

\begin{table}
\scriptsize
\centering
\begin{tabular}{l|c|c|c|c}
         \toprule
         & \multicolumn{4}{c}{Method} \\
             & Rotation & Contrastive & Invariance & Equivariance (Ours)     \\
         \midrule 
         Inference (sec) & 0.016	&0.016	&0.016	&0.016\\
         Detection (sec)	&0.048	&0.049	&0.147	&0.169\\
         Defense (sec)	&0.306	&0.136	&1.616&	1.637 \\
         \bottomrule
    \end{tabular}
    \caption{Running time for different methods on vanilla feedforward inference (Inference), detecting adversarial samples (Detection), and our test-time defense (defense).}
    \vspace{-4mm}
    \label{tab:detect}
\end{table}

% \vspace{-3mm}
\section{Conclusion}
% \vspace{-3mm}
Robust perception under adversarial attacks has been an open challenge. We find that equivariance can be a desired structure to maintain at inference time because it can provide dense structural constraints on a fine-grained level. By dynamically restoring equivariance at inference, we show significant improvement in adversarial robustness across three datasets. Our work hints toward a new direction that uses the right structural information at inference time to improve robustness.

\section*{Acknowledgement}
This research is based on work partially supported by the DARPA SAIL-ON program, the NSF NRI award \#1925157, a GE/DARPA grant, a CAIT grant, and gifts from JP Morgan, DiDi, and Accenture. We thank the anonymous reviewers for their valuable feedback in improving the paper.

\nocite{langley00}

\bibliography{example_paper}
\bibliographystyle{icml2023}

%%%%%%%%%%%%%%%%%%%%%%%%%%%%%%%%%%%%%%%%%%%%%%%%%%%%%%%%%%%%%%%%%%%%%%%%%%%%%%%
%%%%%%%%%%%%%%%%%%%%%%%%%%%%%%%%%%%%%%%%%%%%%%%%%%%%%%%%%%%%%%%%%%%%%%%%%%%%%%%
% APPENDIX
%%%%%%%%%%%%%%%%%%%%%%%%%%%%%%%%%%%%%%%%%%%%%%%%%%%%%%%%%%%%%%%%%%%%%%%%%%%%%%%
%%%%%%%%%%%%%%%%%%%%%%%%%%%%%%%%%%%%%%%%%%%%%%%%%%%%%%%%%%%%%%%%%%%%%%%%%%%%%%%
\newpage
\appendix
\onecolumn
\section{Appendix.}

\subsection{Theoretical Results for Adversarial Robustness}

We now show detailed proof for Lemma 1 and theorem 2.

\begin{lemma}
The standard classifier under adversarial attack is equivalent to predicting with $ P(\Y | \X_a, y^{(s_1)}_a, y^{(s_2)}_a, ..., y^{(s_k)}_a)$, and our approach is equivalent to predicting with $P(\Y | \X_a, y^{(s_1)}, y^{(s_2)}, ..., y^{(s_k)})$.
\end{lemma}

\begin{proof}
For the standard classifier under attack, we know that $P(y^{(s_1)}_a, y^{(s_2)}_a, ..., y^{(s_k)}_a|\X=\x_a) = 1$. Thus we know the standard classifier under adversarial attack is equivalent to
\begingroup\makeatletter\def\f@size{9}\check@mathfonts
\begin{align*}
P(\Y|\X=\x_a) &= \sum_{y^{(s_1)}_a, y^{(s_2)}_a, ..., y^{(s_k)}_a} P(y^{(s_1)}_a, y^{(s_2)}_a, ..., y^{(s_k)}_a | \X=\x_a) P(\Y|y^{(s_1)}_a, y^{(s_2)}_a, ..., y^{(s_k)}_a,\X=x_a)\\
&=P(\Y|y^{(s_1)}_a, y^{(s_2)}_a, ..., y^{(s_k)}_a,\X=\x_a).
\end{align*}
\endgroup
Our algorithm finds a new input image $\x^{(n)}_{\mathrm{max}}$ that
\begingroup\makeatletter\def\f@size{9}\check@mathfonts
\begin{align*}
     &\argmax_{\x^{(n)}} P(\X^{(n)}=\x^{(n)}|\X=\x_a)P(y^{(s_1)}, y^{(s_2)}, ..., y^{(s_k)} |\X^{(n)}=\x^{(n)}) \\
     &= \argmax_{\x^{(n)}} P(\X^{(x)}=\x^{(n)}| \X=\x_a, y^{(s_1)}, y^{(s_2)}, ..., y^{(s_k)}).
\end{align*}
\endgroup

Our algorithm first estimate $\x^{(n)}_{\mathrm{max}}$ with adversarial image $\x_a$ and self-supervised label $\y^{(s)}$. We then predict the label $\Y$ using our new image $\x^{(n)}_{\mathrm{max}}$. Thus, our approach in fact
 estimates $P(\Y|\X^{(n)} = \x^{(n)}_{\mathrm{max}})P(\X^{(n)} =  \x^{(n)}_{\mathrm{max}} |\X=\x_a,  y^{(s_1)}, y^{(s_2)}, ..., y^{(s_k)})$. Note the following holds:
\begingroup\makeatletter\def\f@size{9}\check@mathfonts
\begin{align}
&P(\Y | \X=\x_a, y^{(s_1)}, y^{(s_2)}, ..., y^{(s_k)})\\
&= \sum_{\x^{(n)}} P(\Y|\x^{(n)})P(\x^{(n)}|\X=\x_a,  y^{(s_1)}, y^{(s_2)}, ..., y^{(s_k)}) \\
&\approx P(\Y|\X^{(n)} = \x^{(n)}_{\mathrm{max}})P(\X^{(n)} =  \x^{(n)}_{\mathrm{max}} |\X=\x_a,  y^{(s_1)}, y^{(s_2)}, ..., y^{(s_k)}).
\end{align}
\endgroup
Thus our approach is equivalent to estimating $P(\Y | \X=\x_a, y^{(s_1)}, y^{(s_2)}, ..., y^{(s_k)})$.
\end{proof}

We use the maximum a posteriori (MAP) estimation $\x^{(n)}_{\mathrm{max}}$ to approximate the sum over $\X^{(n)}$ because: (1) sampling a large number of $\X^{(n)}$ is computationally expensive; (2) our results show that random sampling is ineffective; (3) our MAP estimate naturally produces a denoised image that can be useful for other downstream tasks.

\begin{theorem}\label{thm:boundMTL}
Assume the classifier operates better than chance and instances in the dataset are uniformly distributed over $n$ categories. Let the prediction accuracy bounds be $P(\Y|y^{(s_1)}_a, y^{(s_2)}_a, ..., y^{(s_k)}_a, \X_a) \in [b_0, c_0]$, $P(\Y|y^{(s_1)}, \X_a) \in [b_1, c_1]$,  $P(\Y|y^{(s_1)}, y^{(s_2)}, \X_a) \in [b_2, c_2]$, ..., and $P(\Y|y^{(s_1)}, y^{(s_2)}, ..., y^{(s_k)},\X_a) \in [b_k, c_k]$. If the conditional mutual information $I(\Y;y^{(s_i)}| \X_a) > 0$ and $I(\Y;y^{(s_i)}| \X_a, y^{(s_j)}) > 0$ where $i \neq j$, we have $b_0 \leq b_1 \leq ... \leq b_k$ and $c_0 < c_1 < c_2 < ... < c_k$, which means our approach strictly improves the bound for classification accuracy.
\end{theorem}

\begin{proof}
If $I(\Y;y^{(s_i)}| \X=\x_a) > 0$, and $I(\Y;y^{(s_i)}| \X_a, y^{(s_j)}) > 0$ where $i \neq j$, then it is straight-forward that:
\begin{align*}
    I(\Y;y^{(s_1)}, y^{(s_2)}, ..., y^{(s_k)}, \X_a) >I(\Y;y^{(s_i)}, \X_a) > I(\Y;y^{(s_i)}_a, \X_a) = I(\Y; \X_a).
\end{align*}
\begin{align*}
    I(\Y;y^{(s_1)}, y^{(s_2)}, ..., y^{(s_k)}, \X_a) > I(\Y;y^{(s_1)}_a, y^{(s_2)}_a, ..., y^{(s_k)}_a, \X_a) = I(\Y; \X_a).
\end{align*}

We let the predicted label be $\hat{\Y}$, we assume there are $n$ categories, and let the lower bound for prediction accuracy to be $Pr(\hat{\Y}=\Y) \geq 1 - \epsilon_p$. We define $H(\epsilon_p)=-\epsilon_p log \epsilon_p-(1-\epsilon_p)log(1-\epsilon_p)$.
Using the \emph{Fano's Inequality}, we have 
\begin{equation}
    H(\Y|\X_a) \leq H(\epsilon_p) + \epsilon_p \cdot \log(n-1)
\end{equation}

% $$.

\begin{equation}
     -\epsilon_p \cdot \log(n-1) \leq H(\epsilon_p) -H(\Y|\X_a) 
\end{equation}

% $$
We add $H(\Y)$ to both side

\begin{equation}
    H(\Y) - \epsilon_p \cdot \log(n-1) \leq H(\epsilon_p) + I(\Y;\X_a)
\end{equation}
because $I(\Y;\X_a) = H(\Y) - H(\Y|\X_a)$. 

Then we get 

\begin{equation}
\label{eq:fano}
    H(\epsilon_p) + \epsilon_p \log(n-1) \geq - I(\Y;\X_a) + H(\Y)
\end{equation}
% , note that $H(Y) = log(n) \approx log(n-1)$, $H(\epsilon_p) \approx 0$ if error rate $\epsilon_p$ is small.

Now we define a new function  $G(\epsilon_p) = H(\epsilon_p) + \epsilon_p log(n-1)$. Given that in the classification task, the number of categories $n \geq 2$. We know $ log(n-1) \geq 0$. Given that the entropy function $H(\epsilon_p)$ first increase and then decrease, the function $G(\epsilon_p)$ should also first increase, peak at some point, and then decrease.

We calculate the $\epsilon_p$ for the peak value via calculating the first order derivative $G'(\epsilon_p)=0$. By solving this, we have:

\begin{equation}
    \epsilon_p = 1 - \frac{1}{n}
\end{equation}
which shows that the function $G(\epsilon_p)$ is monotonically increasing when $\epsilon_p \in [0, 1 - \frac{1}{n}]$. 

Given that we know, the base classifier already achieves accuracy better than random guessing, thus the given classifier satisfies $\epsilon_p \in [0, 1 - \frac{1}{n}]$.

Now, the function $G(\epsilon_p) = H(\epsilon_p) + \epsilon_p log(n-1)$ is a monotonically increasing function in our studied region, which has the inverse function $G^{-1}$.

By rewritting the equation \ref{eq:fano} We then have 
\begin{equation}
    G(\epsilon_p) \geq - I(\Y;\X_a) + H(\Y)
\end{equation}

We apply the inverse function $G^{-1}$ to both side:
\begin{equation}
    \epsilon_p \geq G^{-1}(- I(\Y;\X_a) + H(\Y))
\end{equation}

\begin{equation}
    1- \epsilon_p \leq 1 - G^{-1}(- I(\Y;\X_a) + H(\Y))
\end{equation}

Note that $(1-\epsilon_p)$ is our defined accuracy. Similarly, we have:

\begin{align*}
1- \epsilon_p &\leq c_1 =  1 - Q^{-1}(- I(\Y;\X_a) + H(\Y)), \\
 1- \epsilon_p &\leq  c_2 =  1 - Q^{-1}(- I(\Y;y^{(s)}, \X_a) + H(\Y)),  \\
 ...\\
 1- \epsilon_p &\leq  c_k =  1 - Q^{-1}(- I(\Y;y^{(s_1)}, y^{(s_2)}, ..., y^{(s_k)}, \X_a) + H(\Y)), 
\end{align*}
where the upper bound is a function of the mutual information. Since $H(\Y)$ is a constant, a larger mutual information will strictly increase the bound. Thus, $c_0 < c_1 < c_2 < ... < c_k$.

In addition, the lower bound will not get worse given the additional information. Thus $b_0 \leq b_1 \leq ... \leq b_k$ and.

\end{proof}

\subsection{Detection}

\textbf{Anomaly detection}, also referred to as novelty detection or outlier detection, predicts the data when the model is uncertain about a deviated model. \cite{ruff2018deep} conducts anomaly detection by training a binary classifier on in-distribution and collected out-of-distribution data, however, it is hard to foresee the out-of-distribution data.
\cite{hendrycks2019using, gidaris2018unsupervised, tack2020csi} need to train the model with self-supervision first and then perform OOD detection using the performance of the self-supervision task. 
In this paper, we will focus on the training-free method that uses sensitivity to estimate the uncertainty of the model~\cite{TFUncertainty}.

\subsubsection{Equivariance for Anomaly Detection}

Let $\y$ be the ground-truth category labels for $\x$. Let the network that uses the feature $\h$ for final task prediction to be $C_{\theta'}$. For prediction, neural networks learn to predict the category $\hat{\y}=C_{\theta'} \circ F_{\theta}(\x) $
by minimizing the loss $L(\hat{\y}, \y)$ between the predictions and the ground truth. For example, for semantic segmentation, $L$ is cross-entropy for each pixel output; for depth prediction, $L$ is an L1 loss for each depth pixel prediction. We define the loss for the final task as follows:
\begin{equation}\label{eq:loss_task}
    \mathcal{L}_t(\x, \y) = L \left(C_{\theta'} \circ F_{\theta}(\x), \y\right),
\end{equation}

As shown in Figure~\ref{fig:city}, \ref{fig:voc}, and \ref{fig:coco}, when the model is uncertain about the input and makes the wrong prediction, it is often \emph{less equivariant}. Work~\cite{tack2020csi, hendrycks2019using} has shown that self-supervision tasks perform worse when the model is uncertain. Thus, we propose to use the equivariance of the output $\hat{\y}=C_{\theta'} \circ F_{\theta}(\x) $ for anomaly detection. We calculate the variance of the output after transformations $g_i$:
\begin{equation}\label{eq:ano}
    \mathcal{L}_{equi}^{output} = \sum_{i} ||g_i^{-1} \circ C_{\theta'} \circ F_{\theta} \circ g_i(\x) - C_{\theta'} \circ F_{\theta}(\x)||^2,
\end{equation}
where we use $C_{\theta'} \circ F_{\theta}(\x)$ as the surrogate mean prediction. Here larger variance indicates less equivariance and therefore higher probability that $\x$ is an out-of-sample data point (see details in~\ref{sec:theory_ano}).

\subsubsection{Theoretical Results for Anomaly detection with Multiple Constraints}\label{sec:theory_ano}

Below we provide theoretical analysis on why the equivariant loss can be used for anomaly detection. 
For each pixel in an image, we denote as $X$ and $Y$ the input pixel and target label. We use $Z_0=\tilde{F}_{\theta}(X)$ and $Z=\tilde{g}^{-1} \circ \tilde{F}_{\theta} \circ \tilde{g}(X)$ to denote the model predictions of the original and transformed input, respectively, where $\tilde{g}$ and $\tilde{F}_{\theta}$ are the associated pixel operations for $g$ and $F_{\theta}$. Correspondingly $e=|Z_0-Y|$ is the error of the model's prediction for the pixel. %Here $e$ can be seen as an indicator of the model's uncertainty: a larger prediction error means that the model should have been less certain at this input. 
Note that $\tilde{g}(\cdot)$ and $\tilde{g}^{-1}(\cdot)$ are equivariant transformations. There can be multiple equivariant transformations $\tilde{g}_i(\cdot)$ for the same input pixel $X$, leading to different model predictions. 
For the same input pixel $X$, we then denote as $\mu(X)=\EB_{\tilde{g}}[Z|X]$ and $\sigma(X)^2=\VB_{\tilde{g}}[Z|X]$ the mean and variance of the model predictions over different equivariant transformations. Here, $\sigma(X)$ measures the \emph{sensitivity} of the model for the input pixel $X$; below we use a shorthand $\sigma$ for $\sigma(X)$ when the context is clear. Following~\cite{TFUncertainty}, we now introduce our model-agnostic assumptions below.

\begin{assumption}[\textbf{Heterogeneous Perturbation}]%\label{assu:heter}
$\epsilon_1 = \frac{Z_0 - \mu(X)}{\sigma(X)} \sim \NM(0,1)$. That is, the model prediction given the original input $X$ behaves like a random Gaussian draw from the model predictions produced by different equivariant transformations. 
\end{assumption}

\begin{assumption}[\textbf{Random Bias}]\label{assu:random}
$\epsilon_2 = Y - \mu(X) \sim \NM(0,B^2)$. That is, the bias of the model prediction behaves like Gaussian noise with bounded variance $B^2$.
\end{assumption}

Assuming each image contains $n$ input pixels, $\{X_i\}_{i=1}^n$, we have the corresponding target labels $\{Y_i\}_{i=1}^n$, errors $\{e_i\}_{i=1}^n$, and sensitivity $\{\sigma_i\}_{i=1}^n$. We denote as $\bar{e}=\frac{1}{n} \sum_i e_i$ the average error of an image. As is usually the case, we further assume the errors are bounded, i.e., $a\leq e_i\leq b$. Our goal is to bound the average pixel error $\bar{e}$ for an image using the image's computed uncertainty (sensitivity) score $\{\sigma_i\}_{i=1}^n$. 
With the assumptions above, % and following~\cite{TFUncertainty}, 
we have:

\begin{theorem}[Estimator for $\bar{e}$]\label{thm:prob_bound}
With probability at least $\delta$, one can estimate the average error $\bar{e}$ for an image using $\sqrt{\frac{2}{\pi}} \EB[\sigma_B]$ with the following guarantee: 
\begin{align*}
|\bar{e} - \sqrt{\frac{2}{\pi}} \EB[\sigma_B]| < \frac{b-a}{\sqrt{n}}\sqrt{\ln \frac{1}{\delta}},
\end{align*}
where $\sigma_B\triangleq \sqrt{\sigma^2+B^2}$ is the smoothed version of the uncertainty (sensitivity) $\sigma$ and $B$ is a constant from \asuref{assu:random}. 
\end{theorem}

\textbf{Estimated Average Error for Anomaly Detection.} We can see that $\sqrt{\frac{2}{\pi}} \EB[\sigma_B]$ can be a good estimate for the average error for an image, and that this estimate gets more accurate as $n$ gets larger. Therefore, it can be used directly for anomaly detection; larger $\sqrt{\frac{2}{\pi}} \EB[\sigma_B]$ indicates a potentially larger error, meaning that the image is more likely to be an anomaly. Note that the expectation $\EB[\sigma_B]$ is over the space of pixels in all images governed by the assumptions above, not over the pixels in a specific image. In practice, we estimate $\EB[\sigma_B]$ by averaging over the pixel-level sensitivity in an image, i.e., $\frac{1}{n}\sum_i \sqrt{\sigma^2_i+B^2}$, leading to \eqnref{eq:ano}. 

\textbf{Extension to the Multivariate Case.} \thmref{thm:prob_bound} assumes one scalar output for \emph{each pixel} in an image; this is directly applicable for dense regression tasks, e.g., depth estimation. For dense classification tasks, e.g., segmentation, the label for each pixel is represented by a one-hot vector. Fortunately, \thmref{thm:prob_bound} can be naturally extended to the multivariate case, and therefore works for both regression and classification tasks. Note that in classification tasks,  $\mu(X)$ and $\sigma(X)$ are both real-valued vectors where all entries in a vector sums up to $1$, while $Y$ and $Z_0$ are both one-hot vectors (vectors with one entry equal to $1$ and others equal to $0$); therefore the two mild assumptions above are still reasonable.

\begin{figure}[t]
\centering
\includegraphics[width=0.85\textwidth]{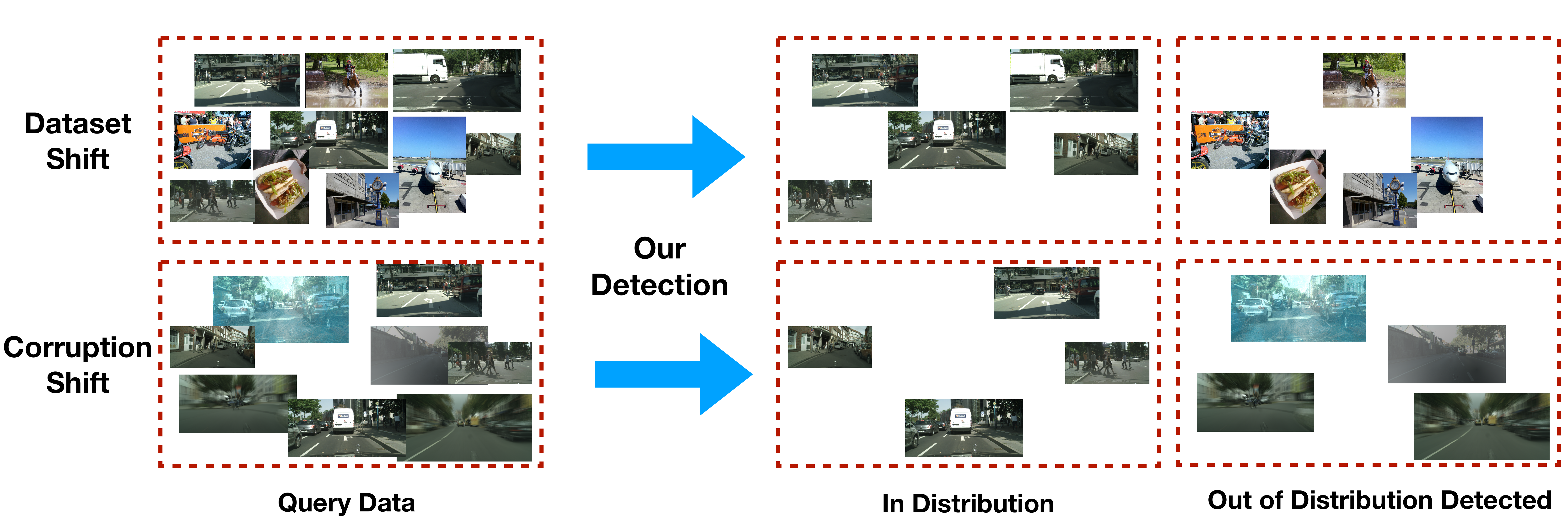}
\vspace{-3mm}
  \caption{\small{Using equivariance can detect both dataset shifting and corruption shifting.}}
  \label{fig:detection}
  \vspace{-5mm}
\end{figure}

\subsubsection{Robustness on Anomaly Detection}
\begin{table}[t]
\centering
\setlength{\tabcolsep}{0.8pt}
\scriptsize
\caption{\small{AUROC (multiplied by 100) of anomaly detection on corrupted images. Our equivariance method achieves better detection efficiency over 15 types of corruptions.
%~\citep{imgnet-C} set with 15 different corruptions. While ViT models are more robust than the state-of-the-art convolution models, our method further reduces the mCE by over 4\%.
}}
% \vspace{-3mm}
    \begin{tabular}{l|ccc|cccc|cccc|cccc}
         \toprule
         & \multicolumn{15}{c}{Cityscape} \\
          Model  & Gauss & Shot & Impul & Defoc & Glass & Motion & Zoom & Snow & Frost & Fog & Bright & Cont & Elast & Pixel & JPEG \\
         \midrule 
         Rot~\cite{hendrycks2019using} & 57 & 54 & 49 & 43 & 55 & 35 & 44 & 39 & 64 & 51 & 46 & 52 & 42 & 44 & 59\\
         CSI~\cite{tack2020csi} & 67 & 67 & 62 & 65 & 62 & 57 & 64 & 45 & 55 & 63 & 63 & 65 & 55 & 53 & 54 \\
         Inv &  99 & 99 & 99 & 100 & 94 & 87 & 80 & 86 & 95 & 93 & 86 & 98 & 52 & 98 & 100\\
        %  Inv (discrete) & \\
         Ours &\textbf{100} & \textbf{100} & \textbf{100} & \textbf{100} & \textbf{99} &\textbf{98} & \textbf{99} & \textbf{100} & \textbf{100} & \textbf{98} & \textbf{94} & \textbf{100} & \textbf{77} & \textbf{99} & \textbf{100}\\
         \bottomrule
         \toprule
         & \multicolumn{15}{c}{PASCAL VOC} \\
          Model  & Gauss & Shot & Impul & Defoc & Glass & Motion & Zoom & Snow & Frost & Fog & Bright & Cont & Elast & Pixel & JPEG \\
         \midrule 
         Rot~\cite{hendrycks2019using} & 37 & 39 & 39 & 55 & 53 & 54 & 54 & 43 & 49 & 55 & 51 & 55 & 50 & 52 & 50 \\
         CSI~\cite{tack2020csi} & 49 & 51 & 50 & 55 & 55 & 58 & 54 & 61 & 58 & 58 & 52 & 72 & 53 & 54 & 54 \\
         Inv & \textbf{99} & \textbf{99} & \textbf{99} & 66 & 21 & 36 & 37 & 70 & 74 & 68 & 54 & 66 & 45 & 35 & 40 \\
        %  Inv (discrete) & \\
         Ours & 98 & 98 & 98 & \textbf{96} & \textbf{95} & \textbf{91} & \textbf{93} & \textbf{85} & \textbf{86} & \textbf{81} & \textbf{60} & \textbf{92} & \textbf{70} & \textbf{85} & \textbf{75} \\
         \bottomrule
         \toprule
         & \multicolumn{15}{c}{MSCOCO} \\
          Model  & Gauss & Shot & Impul & Defoc & Glass & Motion & Zoom & Snow & Frost & Fog & Bright & Cont & Elast & Pixel & JPEG \\
         \midrule 
         Rot~\cite{hendrycks2019using} & 95 & 95 & 95 & 93 & 93 & 93 & 93 & 93 & 94 & 93 & 93 & 93 & 93 & 93 & 92\\
         CSI~\cite{tack2020csi} & 88 & 89 & 86 & 76 & 79 & 77 & 75 & 42 & 47 & 44 & 77 & 22 & 84 & 82 & 85\\
         Inv & 98 & 98 & 98 & 96 & 97 & 96 & 96 & 98 & 98 & 97 & 98 & 98 & 97 & 97 & 97 \\
        %  Inv (discrete) & \\
         Ours & \textbf{98} & \textbf{98} & \textbf{98} & \textbf{98} & \textbf{98} & \textbf{98} & \textbf{98} & \textbf{98} & \textbf{98} & \textbf{98} & \textbf{98} & \textbf{98} & \textbf{98} & \textbf{98} & \textbf{98}\\
         \bottomrule
         
    \end{tabular}
% \vspace{-2mm}
\label{tab:city-C}
\end{table}

\begin{SCtable}
\scriptsize
\begin{tabular}{l|ccc|ccc|ccc|ccc}
         \toprule
         & \multicolumn{3}{c}{Rotation} & \multicolumn{3}{c}{Contrast} & \multicolumn{3}{c}{Invariance} & \multicolumn{3}{c}{Equivariance}\\
          Model  & CI  & VO & CO & CI  & VO & CO & CI  & VO & CO & CI  & VO & CO\\
         \midrule 
         CI & - & 56 & 51 & - & 72 & 52 & - & 93 & 91 & - & \textbf{98} &\textbf{ 91} \\
         VO & 67 & - & 61 & 54 & - & 85 & \textbf{95} & - & \textbf{91} & 71 & - & 88 \\
         CO & \textbf{72} & 93 & - & 69 & 83 & - & 51 & 96 & - & 55 & \textbf{99} & -  \\
         \bottomrule
    \end{tabular}
    \caption{\small{AUROC for out of dstribution detection. The rows are the source data that the models have been trained on. The columns are the data where the OOD are sampled from.}}
    \label{tab:detect-dataset}
\end{SCtable}

% TODO: OOD across dataset? Discuss with Lingyu,

\textbf{Dataset and Tasks.} We conduct anomaly detection experiments with 15 common corruptions~\cite{imgnet-C} on Cityscapes, PASCAL VOC, and MSCOCO. We also study whether using equivariance can detect examples from a different dataset.

\noindent \textbf{Baselines.} CSI \cite{tack2020csi} uses contrastive loss as indicator for novelty detection. Hendrycks et al.~\cite{hendrycks2019using} (Rot) uses rotation task's performance for detection. Invariance (Inv) uses the consistency between different views of the same image for anomaly detection, which uses the same setup as our equivariance method except for the reversed transformation $g^{-1}$ in feature space.

\noindent \textbf{Results.} We show visualization of the task in Figure~\ref{fig:detection}. Table~\ref{tab:city-C} shows the detection performance on \emph{corrupted images}. Our approach in general improves the AUROC score over the baselines, achieving up to 15 AUROC points improvement, which demonstrates the corruption detection of our approach. 
 We show results on detecting \emph{dataset shifting} in Table~\ref{tab:detect-dataset}. We denote Cityscapes, PASCAL VOC, and COCO as CI, VO, and CO, respectively. Each row indicates the source model that we trained on, and each column is the OOD sampled to detect. Our method in general achieves better out-of-distribution detection efficiency over the existing approaches.

 \subsubsection{Theoretical Results for Anomaly Detection}
In the main paper, we provide theoretical analysis on why the equivariant loss can be used for anomaly detection. We now show detailed proof for our Theorem 1.

\begin{theorem}[Estimator for $\bar{e}$]\label{thm:prob_bound}
With probability at least $\delta$, one can estimate the average error $\bar{e}$ for an image using $\sqrt{\frac{2}{\pi}} \EB[\sigma_B]$ with the following guarantee: 
\begin{align*}
|\bar{e} - \sqrt{\frac{2}{\pi}} \EB[\sigma_B]| < \frac{b-a}{\sqrt{n}}\sqrt{\ln \frac{1}{\delta}},
\end{align*}
where $\sigma_B\triangleq \sqrt{\sigma^2+B^2}$. 
\end{theorem}

\begin{proof}
By the law of total expectation, we have
\begin{align*}
\EB [e] &= \EB_{\sigma} \EB[e | \sigma] 
= \EB_{\sigma} \EB\left[|\sigma \epsilon_1 - \epsilon_2| | \sigma \right]
\\&= \EB_{\sigma} \EB\left[|\NM(0, \sigma^2+B^2)| | \sigma \right]
\\&= \sqrt{\frac{2}{\pi}} \EB_{\sigma}[\sqrt{\sigma^2+B^2}] % ?
\\&\triangleq \sqrt{\frac{2}{\pi}} \EB[\sigma_B],
\end{align*}
where $\sigma_B\triangleq \sqrt{\sigma^2+B^2}$. Defining the total error for an image of $n$ pixels $S_n=\sum_{i=1}^n e_i$ and by Hoeffding's inequality, we then have
\begin{align}
P(|S_n-\EB[S_n]|\geq t) \leq \exp(-\frac{t^2}{n(b-a)^2}),\label{eq:hoeffding}
\end{align}
where 
\begin{align}
\EB[S_n]=\sum\nolimits_{i=1}^n \EB[e_i] = n \sqrt{\frac{2}{\pi}} \EB[\sigma_B]. \label{eq:es}
\end{align}
Combining \eqnref{eq:es} and \eqnref{eq:hoeffding}, we have
\begin{align*}
P(|\bar{e}-\sqrt{\frac{2}{\pi}} \EB[\sigma_B]|\geq \frac{t}{n}) \leq \exp(-\frac{t^2}{n(b-a)^2}),
\end{align*}
where $\bar{e}$ is the average error of an image. Setting $\delta = \exp (-\frac{t^2}{n(b-a)^2})$, we then have that with probability at least $\delta$, 
\begin{align*}
|\bar{e} - \sqrt{\frac{2}{\pi}} \EB[\sigma_B]| < \frac{b-a}{\sqrt{n}}\sqrt{\ln \frac{1}{\delta}}.
\end{align*}
\end{proof}

\subsection{Additional Analysis}
\subsubsection{Preserving clean accuracy by detecting first}
\label{subsubsec:detect}
In deploying our defense, one can further preserve the accuracy on clean images by a detect-then-defend algorithm described below. Based on our findings that clean images and attacked images have a large difference in the average equivariance score (Table~\ref{tab:equi_metric}), we can set a threshold value to determine whether or not deploy our defense for a potentially attacked image based on its equivariance score. Experimental results are reported below in Table~\ref{tab:detect_first}. As shown in the table, clean accuracy can be preserved to a large degree without significant reduction in defense performance.

\begin{table}
\scriptsize
\centering
\begin{tabular}{l|c|c|c|c|c|c}
         \toprule
         & \multicolumn{6}{c}{Method} \\
             & Vanilla & Random & Rotation & Contrastive & Invariance & Equivariance (Ours)     \\
         \midrule 
         Cityscape	& 58.29	& 55.33	& 49.82	& 50.85	& 36.94	& 51.13\\
            Pascal	& 69.52& 	69.24	& 49.07	& 68.22	& 66.58	& 63.05\\
            COCO	& 63.02	& 63.03	& 61.69	& 56.76	& 56.06 & 	58.71\\
            COCO Instance	& 34.5	& 33.6	& 33.3	& 29.8	& 26.7	& 34.5\\
         \bottomrule
    \end{tabular}
    \caption{Clean accuracy by first detecting adversarial example. By avoiding running test-time optimization on clean examples, we can largely improve clean accuracy.}
    \vspace{-4mm}
    \label{tab:detect_first}
\end{table}

\subsubsection{Ablation studies on optimizer}
In the paper, we use the SGLD optimizer which add noise during optimization. We compare the performance of using SGLD and SGD optimizer without noise for our defense in the Table~\ref{tab:optimizer} below. While our method is both effective with both optimization algorithm, SGLD achieves higher robustness.

\begin{table}[h]
\scriptsize
\centering
\begin{tabular}{l|c|c}
         \toprule
         & \multicolumn{2}{c}{MS COCO} \\
             &  Equivariance SGLD & Equivariance SGD\\
         \midrule 
         PGD & \textbf{24.51} & 16.68\\
         \bottomrule
    \end{tabular}
    \caption{{Effect of optimizer}}
    \label{tab:optimizer}
\end{table}

\subsubsection{Runtime analysis}
We report inference speed of our defense in Table~\ref{tab:runtime}. It is worth noting that if we deploy detection first, as mentioned in section~\ref{subsubsec:detect}, the defense will skip the majority of clean images and will not have a large effect on runtime. For attacked images, our algorithm is 40 times slower due to the test-time optimization. Given the rare cases of adversarial examples, this delay is reasonable. We can spend more time on the hard adversarial examples, as there is no point of making the wrong predictions only to make it fast.

\begin{table}[h]
\scriptsize
\centering
\begin{tabular}{l|c|c}
         \toprule
         & \multicolumn{2}{c}{MS COCO} \\
             &  Vanilla & Equivariance\\
         \midrule 
         runtime (s) & 0.046 & 1.699\\
         \bottomrule
    \end{tabular}
    \caption{{Runtime analysis. Results are measured on a single A6000 GPU, averaged across 100 examples.}}
    \label{tab:runtime}
\end{table}
 
 \section{Visualization}

\begin{figure}[]
\centering
\includegraphics[width=\textwidth]{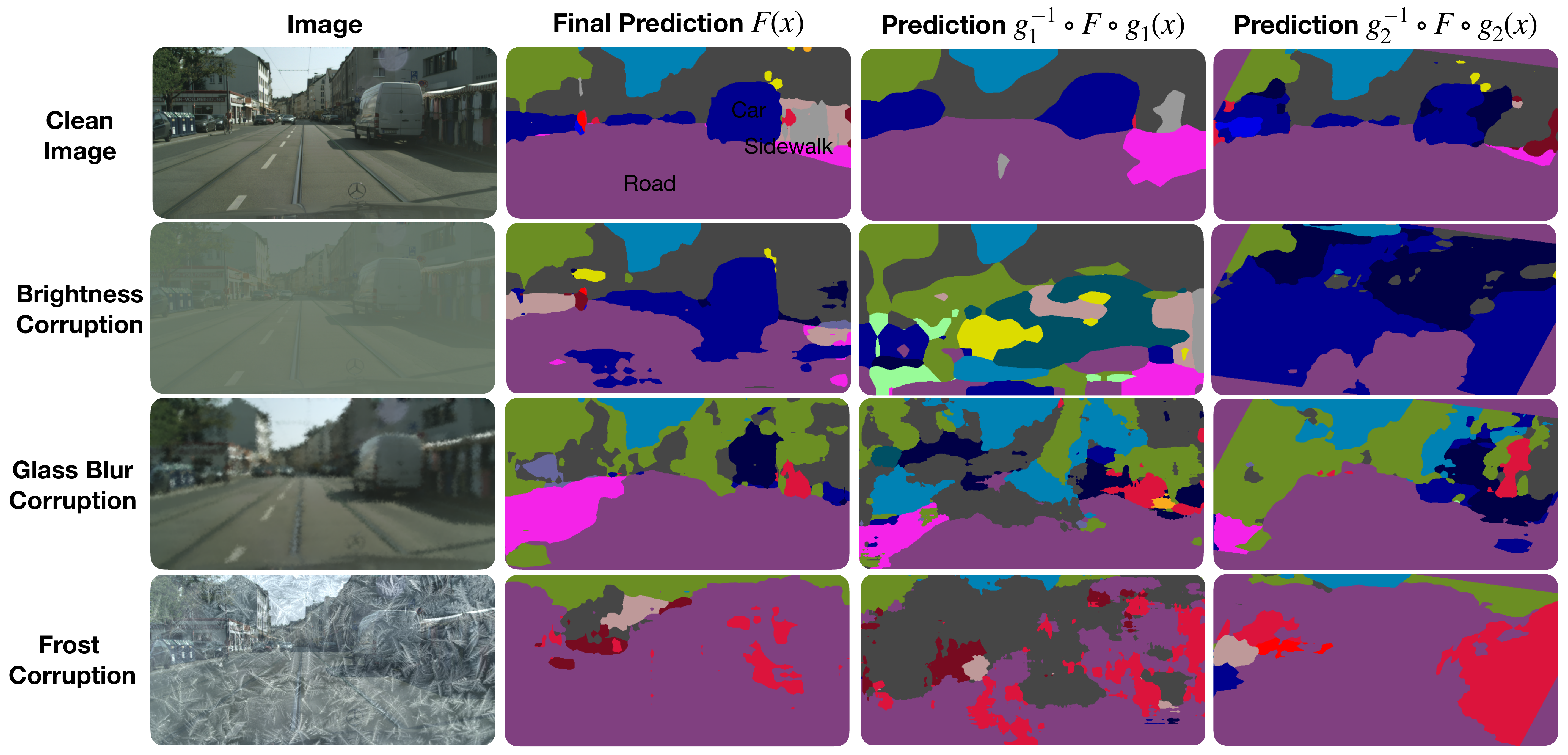}
  \caption{{Examples showing equivariance on clean image and corrupted images on Cityscapes dataset. Clean image is equivariant. Images under corruption are not equivariant, allowing us to detect corruption using equivariance measurement.}}
  \label{fig:city}
\end{figure}

\begin{figure}[]
\centering
\includegraphics[width=\textwidth]{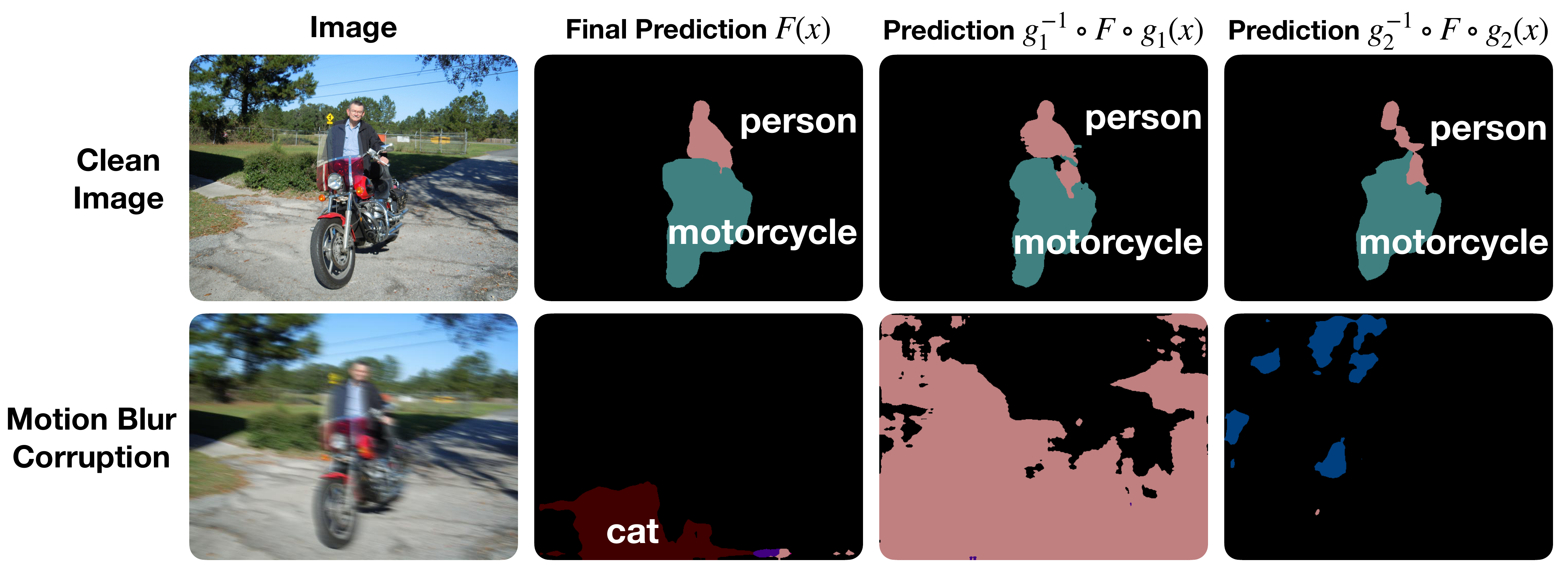}
  \caption{{Examples showing equivariance on clean image and a random corrupted image on COCO dataset. Clean image is equivariant. Images under corruption are not equivariant, allowing us to detect corruption using equivariance measurement.}}
  \label{fig:coco}
\end{figure}

We show additional visualization on the equivarance of representation when the input suffers from natural corruptions. In Figure~\ref{fig:city} and \ref{fig:coco},we show random visualizations on Cityscapes and PASCAL VOC.

%%%%%%%%%%%%%%%%%%%%%%%%%%%%%%%%%%%%%%%%%%%%%%%%%%%%%%%%%%%%%%%%%%%%%%%%%%%%%%%
%%%%%%%%%%%%%%%%%%%%%%%%%%%%%%%%%%%%%%%%%%%%%%%%%%%%%%%%%%%%%%%%%%%%%%%%%%%%%%%

\end{document}